\documentclass[10pt,twocolumn,letterpaper]{article}

\usepackage{iccv}
\usepackage{times}
\usepackage{epsfig}
\usepackage{graphicx}
\usepackage{amsmath}
\usepackage{amssymb}
\usepackage{color,soul}
\usepackage{xspace}
\usepackage{caption}
\usepackage{pdfpages}
\usepackage{color}
\usepackage{enumitem}
\setlist{nolistsep}
\usepackage{makecell}

\newsavebox\dotbox
\sbox{\dotbox}{\(\displaystyle\bigodot\)}

\usepackage{gensymb}
\usepackage{amsfonts}
\usepackage{multirow}
\usepackage[numbers]{natbib}

\usepackage{algorithm}% http://ctan.org/pkg/algorithms
\usepackage[noend]{algpseudocode}
\newcommand{\tabincell}[2]{\begin{tabular}{@{}#1@{}}#2\end{tabular}}
\usepackage[pagebackref=true,breaklinks=true,letterpaper=true,colorlinks,bookmarks=false]{hyperref}
\newcommand{\gantree}{\textit{GAN-Tree}\xspace}
\newcommand{\igantree}{\textit{iGAN-Tree}\xspace}
\newcommand{\ganset}{\textit{GAN-Set}\xspace}
\newcommand{\gnode}{\textit{GN}\xspace}

\newcommand{\norm}[1]{\left\lVert#1\right\rVert}
\newcommand{\encoder}[2]{$E^{(#1)}_{#2}$}
\newcommand{\decoder}[1]{$G^{(#1)}$}
\newcommand{\disc}[1]{$D^{(#1)}$}

\iccvfinalcopy % *** Uncomment this line for the final submission

\def\iccvPaperID{3932} % *** Enter the CVPR Paper ID here

\ificcvfinal\fi
\begin{document}
%\pagenumbering{gobble}

%%%%%%%%% TITLE
\title{GAN-Tree: An Incrementally Learned Hierarchical Generative Framework for Multi-Modal Data Distributions}

\author{Jogendra Nath Kundu\thanks{equal contribution}% - listed in reverse alphabetical order of last names} 
\qquad Maharshi Gor\footnotemark[1] \qquad Dakshit Agrawal \qquad R. Venkatesh Babu\\
%Video Analytics Lab, Department of Computational and Data Sciences\\
Video Analytics Lab, Indian Institute of Science, Bangalore, India\\
{\tt\small jogendrak@iisc.ac.in, maharshigor18@gmail.com, dagrawal@cs.iitr.ac.in, venky@iisc.ac.in} 
%\{krishnaphaniiitg, anujpahuja13\}@gmail.com, venky@iisc.ac.in}
%\and
}

\maketitle
\thispagestyle{empty}

%%%%%%%%% ABSTRACT
% \begin{abstract}
% To alleviate the problem of modeling multi-modal data distributions, techniques with multi-mode prior or multiple generator models have been proposed. But such approaches do not provide flexibility to dynamically increase or decrease the number of modes or the number of generators in prior, due to which they may fail depending on the empirically chosen initial mode components. In contrast, the proposed iGAN-Tree framework follows a top-down hierarchical divisive strategy to address such discontinuous multi-modal data. Making no assumptions on the choice of the number of modes, iGAN-Tree utilizes a novel mode-splitting algorithm to effectively split the parent mode to semantically cohesive children modes, at the same time facilitating unsupervised clustering. Further, it also enables tal training and addition of more modes to an already trained Gan-Tree network, by updating only a single branch of the tree structure. As compared to prior State-of-the art approaches, the proposed framework offers a higher degree of flexibility in choosing a large variety of certain sets of mutually exclusive and exhaustive nodes called GAN-Set, which enables generation of novel samples according to the requirement of quality-vs-diversity trade-off.
% \end{abstract}

\begin{abstract}
Despite the remarkable success of generative adversarial networks, their performance seems less impressive for diverse training sets, requiring learning of discontinuous mapping functions. Though multi-mode prior or multi-generator models have been proposed to alleviate this problem, such approaches may fail depending on the empirically chosen initial mode components. In contrast to such bottom-up approaches, we present GAN-Tree\footnote{Code available at \href{https://github.com/val-iisc/GANTree}{https://github.com/val-iisc/GANTree}}, which follows a hierarchical divisive strategy to address such discontinuous multi-modal data. Devoid of any assumption on the number of modes, GAN-Tree utilizes a novel mode-splitting algorithm to effectively split the parent mode to semantically cohesive children modes, facilitating unsupervised clustering. Further, it also enables incremental addition of new data modes to an already trained GAN-Tree, by updating only a single branch of the tree structure. As compared to prior approaches, the proposed framework offers a higher degree of flexibility in choosing a large variety of mutually exclusive and exhaustive tree nodes called GAN-Set. Extensive experiments on synthetic and natural image datasets including ImageNet demonstrate the superiority of GAN-Tree against the prior state-of-the-art. %Code available at \href{https://github.com/val-iisc/GANTree}{this link}.

%in achieving state-of-the-art performance over prior art.
\end{abstract}

%%%%%%%%% BODY TEXT
\section{Introduction}
\label{section:intro}

Generative models have gained enormous attention in recent years as an emerging field of research to understand and represent the vast amount of data surrounding us. The primary objective behind such models is to effectively capture the underlying data distribution from a set of given samples. The task becomes more challenging for complex high-dimensional target samples such as image and text. %However, 
%%Emergence of deep generative models has achieved considerable progress in this direction by effectively formalizing the objective function, which enables such models to be trained using the back-propagation procedure. 
Well-known techniques like Generative Adversarial Network (GAN)~\cite{goodfellow2014generative} and Variational Autoencoder (VAE)~\cite{kingma2013auto} realize it by defining a mapping from a predefined latent prior to the high-dimensional target distribution.

%%%%%%%%%%%%%%%%%%%%%%%% Figure 1a %%%%%%%%%%%%%%%%%%%%%
\begin{figure}[t]%[h!]
\centering    
	\includegraphics[width=0.93\linewidth]{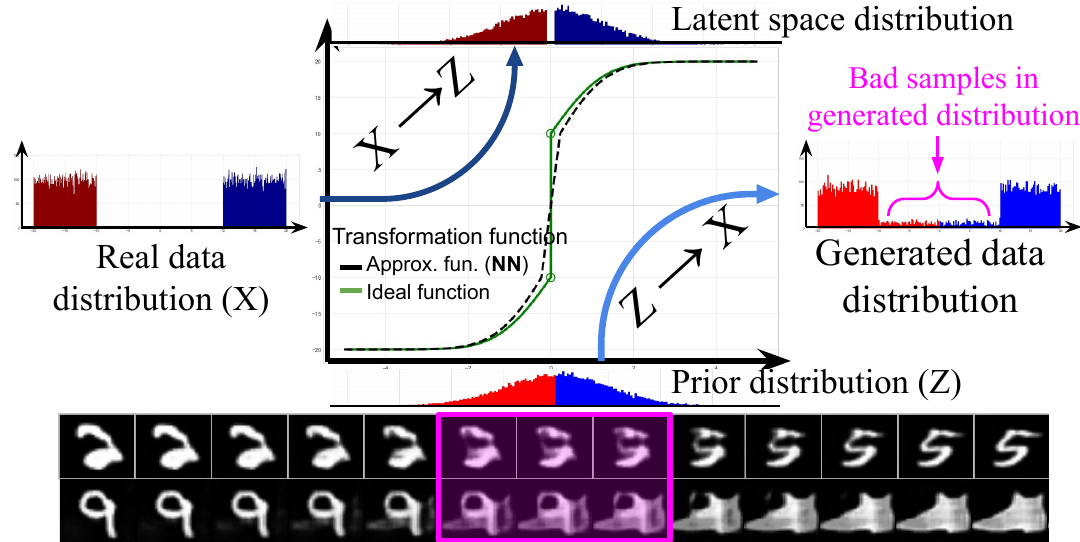}%cfig_4_final_copy
	\vspace{-5pt}
	\caption{%\fontsize{7}{1}\selectfont 
	Illustration of an ideal mapping (green plot, a non-invertible mapping of a disconnected uniform distribution to a uni-modal Gaussian), and its invertible approximation (dotted plot) learned by a neural network. The approximate mapping ($X\rightarrow Z$) introduces a discontinuity in the latent-space (top), whose inverse ($Z\rightarrow X$) when used for generation from a uni-modal prior (bottom) reveals implausible samples (in purple).
	%\\ generation of implausible samples (highlighted in purple).\vspace{-14pt}
	%Using the learned approximate function for transforming real data distribution (left) to its latent representation introduces a discontinuity in latent space distribution (top), whose inverse when used for generating real data samples from a prior uni-modal distribution (bottom) reveals generation of bad samples indicated in purple box.
	\vspace{-12pt}
	}
	\label{fig:concept}  
\end{figure}
%%%%%%%%%%%%%%%%%%%%%%%% Figure 1a ends %%%%%%%%%%%%%%%%%

Despite the success of GAN, the potential of such a framework has certain limitations. GAN is trained to look for the best possible approximate $P_g(X)$ of the target data distribution $P_d(X)$ within the boundaries restricted by the choice of latent variable setting (i.e. the dimension of latent embedding and the type of prior distribution) and the computational capacity of the generator network (characterized by its architecture and parameter size). Such a limitation is more prominent in the presence of highly diverse intra-class and {inter-class} variations, where the given target data spans a highly sparse non-linear manifold. This indicates that the underlying data distribution $P_d(X)$ would constitute multiple, sparsely spread, low-density regions. Considering enough capacity of the generator architecture (Universal Approximation Theorem~\cite{hornik1989multilayer}), GAN guarantees convergence to the true data distribution. However, the validity of the theorem does not hold for mapping functions involving discontinuities (Fig.~\ref{fig:concept}), as exhibited by natural image or text datasets. Furthermore, various regularizations~\cite{che2016mode,salimans2016improved} imposed in the training objective inevitably restrict the generator to exploit its full computational potential.

A reasonable solution to address the above limitations could be to realize multi-modal prior in place of the single-mode distribution in the general GAN framework. Several recent approaches explored this direction by explicitly enforcing the generator to capture diverse multi-modal target distribution~\cite{gurumurthy2017deligan,khayatkhoei2018disconnected}.
The prime challenge encountered by such approaches is attributed to the choice of the number of modes to be considered for a given set of fully-unlabelled data samples. To better analyze the challenging scenario, let us consider an extreme case, where a very high number of modes is chosen in the beginning without any knowledge of the inherent number of categories present in a dataset. In such a case, the corresponding generative model would deliver a higher inception score~\cite{barratt2018note} as a result of dedicated prior modes for individual sub-categories or even sample level hierarchy. This is a clear manifestation of “\textit{overfitting in generative modeling}” as such a model would generate reduced or a negligible amount of novel samples as compared to a single-mode GAN. Intuitively, the ability to interpolate between two samples in the latent embedding space \cite{zhu2016generative,radford2015unsupervised} demonstrates continuity and generalizability of a generative model. However, such an interpolation is possible only within a pair of samples belonging to the same mode specifically in the case of multi-modal latent distribution. It reveals a clear trade-off between the two schools of thoughts, that is, multi-modal latent distribution has the potential to model a better estimate of $P_d(X)$ as compared to a single-mode counterpart, but at a cost of reduced generalizability depending on the choice of mode selection. This also highlights the inherent trade-off between quality (multi-modal GAN) and diversity (single-mode GAN) of a generative model \cite{nguyen2017plug} specifically in the absence of a concrete definition of natural data distribution. %Hence there is a strong incentive to design frameworks or algorithms with enough flexibility to explore the number of modes required to achieve an effective balance between the above two aspects. 

An ideal generative framework addressing the above concerns must have the following important traits: 
\begin{itemize}
\item The framework should allow enough flexibility in the design choice of the number of modes to be considered for the latent variable distribution. 

\item Flexibility in generation of novel samples depending on varied preferences of quality versus diversity according to the intended application in focus (such as unsupervised clustering, hierarchical classification, nearest neighbor retrieval, etc.).

\item Flexibility to adapt to a similar but different class of additional data samples introduced later in absence of the initial data samples (incremental learning setting). 
\end{itemize}

\label{exclusive}
In this work, we propose a novel generative modeling framework, which is flexible enough to address the quality-diversity trade-off in a given multi-modal data distribution. We introduce \gantree, a hierarchical generative modeling framework consisting of multiple GANs organized in a specific order similar to a binary-tree structure. In contrast to the bottom-up approach incorporated by recent multi-modal GAN~\cite{tolstikhin2017adagan, khayatkhoei2018disconnected, gurumurthy2017deligan}, we follow a top-down hierarchical divisive clustering procedure. First, the root node of the \gantree is trained using a single-mode latent distribution on the full target set aiming maximum level of generalizability. Following this, an unsupervised splitting algorithm is incorporated to cluster the target set samples accessed by the parent node into two different clusters based on the most discriminative semantic feature difference. %Intuitively the splitting algorithm is designed to effectively push the samples to one of the two children modes, which best identifies them. 
After obtaining a clear cluster of target samples, a bi-modal generative training procedure is realized to enable the generation of plausible novel samples from the predefined children latent distributions. 
%To enable inference, the encoder network at each tree-node is utilized as a router to route samples from the root node to one of the leaf nodes which best identifies it. This procedure is further repeated for each child node where the order of \textcolor{red}{the next node to be} split is decided based on the node with minimum likelihood of the corresponding cluster of target samples. 
To demonstrate the flexibility of \gantree, we define \ganset, a set of mutually exclusive and exhaustive tree-nodes which can be utilized together with the corresponding prior distribution to generate samples with the desired level of quality vs diversity. Note that the leaf nodes would realize improved quality with reduced diversity whereas the nodes closer to the root would yield a reciprocal effect. %The extent of flexibility offered by \gantree can be demonstrated by a particular \ganset selection with some nodes close to the root along with some leaf nodes satisfying the mutually exclusive and exhaustiveness constraint. 

%We also introduce an incremental flavour to the proposed framework - \igantree, It supports incremental upgradation, which is highly challenging to realize in general GAN frameworks. Firstly, it allows incremental generative modeling in a much efficient manner as only a certain branch of the full GAN-Tree has to be updated to effectively model distribution of a new input set. Secondly, the top-down setup results in an unsupervised clustering of the underlying class-labels as a byproduct, which can be further utilized to develop a classification model with implicit hierarchical categorization.

The hierarchical top-down framework opens up interesting future upgradation possibilities, which is highly challenging to realize in general GAN settings. One of them being incremental \gantree, denoted as \igantree. It supports incremental generative modeling in a much efficient manner, as only a certain branch of the full \gantree has to be updated to effectively model distribution of a new input set. Additionally, the top-down setup results in an unsupervised clustering of the underlying class-labels as a byproduct, which can be further utilized to develop a classification model with implicit hierarchical categorization.

\section{Related work}
%GAN has been used extensively in variety of different applications like image inpainting~\cite{ledig2017photo}, image translation~\cite{isola2017image}, unsupervised feature learning~\cite{salimans2016improved}, few-short learning etc. %It is used to model unknown data distribution inferred from huge amount of natural real samples $\mathcal{X}$.
Commonly, most of the generative approaches realize the data distribution as a mapping from a predefined prior distribution~\cite{goodfellow2014generative}.
%%($\mathcal{Z}$) called generator network such that $G:\mathcal{Z}\rightarrow\mathcal{X}$~\cite{goodfellow2014generative}. %In case of GAN~\cite{goodfellow2014generative}, a discriminator $D$ is trained to learn the discrepancy between $\mathcal{X}$ and $G(\mathcal{Z})$, whereas the generator $G$ is trained to minimize the discrepancy by fooling the discriminator network in a two player minimax game setup. %Goodfellow \etal~\cite{goodfellow2014generative} have proven that it will converge to the actual data distribution under certain assumptions. 
BEGAN~\cite{berthelot2017began} proposed an autoencoder based GAN, which adversarially minimizes an energy function~\cite{zhao2016energy} derived from Wasserstein distance~\cite{arjovsky2017wasserstein}.
Later, several deficiencies in this approach have been explored, such as mode-collapse~\cite{srivastava2017veegan}, unstable generator convergence~\cite{metz2016unrolled,arjovsky2017towards}, etc. Recently, several approaches propose to use an inference network~\cite{che2016mode,larsen2016autoencoding}, $E:\mathcal{X}\rightarrow\mathcal{Z}$, or minimize the joint distribution $P(\mathcal{X},\mathcal{Z})$~\cite{donahue2016adversarial,dumoulin2016adversarially} to regularize the generator from mode-collapse. %Larsen~\etal~\cite{larsen2016autoencoding} proposed a hybrid variational autoencoder with GAN framework to effectively leverage advantages of both the approaches. 
%Other approaches use the discriminator to minimize the discrepancy between the joint distribution $P(\mathcal{X}, \mathcal{Z})$~\cite{donahue2016adversarial,dumoulin2016adversarially}. %%considering both the generator along with an inference network.
Although these approaches effectively address mode-collapse, they suffer from the limitations of modeling disconnected multi-modal data~\cite{khayatkhoei2018disconnected}, using single-mode prior and the capacity of single generator transformation as discussed in Section~\ref{section:intro}.

To effectively address multi-modal data, two different approaches have been explored in recent works viz. a) multi-generator model and b) single generator with multi-mode prior. Works such as~\cite{ghosh2017multi,hoang2018mgan,khayatkhoei2018disconnected} propose to utilize multiple generators to account for the discontinuous multi-modal natural distribution. These approaches use a mode-classifier network either separately~\cite{hoang2018mgan} or embedded with a discriminator~\cite{ghosh2017multi} to enforce learning of mutually exclusive and exhaustive data modes dedicated to individual generator network. Chen~\etal~\cite{chen2016infogan} proposed Info-GAN, which aims to exploit the semantic latent source of variations by maximizing the mutual information between the generated image and the latent code. Gurumurthy~\etal~\cite{gurumurthy2017deligan} proposed to utilize a Gaussian mixture prior with a fixed number of components in a single generator network. These approaches used a fixed number of Gaussian components and hence do not offer much flexibility on the scale of quality versus diversity required by the end task in focus. Inspired by boosting algorithms, AdaGAN~\cite{tolstikhin2017adagan} proposes an iterative procedure, which incrementally addresses uncovered data modes by introducing new GAN components using the sample re-weighting technique. %A thorough comparison of \gantree against AdaGAN~\cite{tolstikhin2017adagan} and DMGAN~\cite{khayatkhoei2018disconnected} is presented in Section~\ref{section:experiments}.

% \begin{algorithm}[tb]

% \caption{Sampling Algorithm}
% \label{algo:Sampling}
% \begin{algorithmic}[1]
% \State \textbf{input:} class \textit{PDFs}, $P_c(x;\mu_c,\Sigma_c)$, and source \textit{PDF}, $P_s(x;\mu_s; \Sigma_s)$, number of required samples $N$
% \State /* $||$ signifies an Append Operation */
% \State $\hat X_s \gets \{\}$
% \State $\hat Y_s \gets \{\}$
% \While{$|\hat X_s| \leq N$}
%   \State Let $\lambda_{c}, v_{c}$ be the maximum eigen value and
%   \State corresponding eigen vector of $\Sigma_c$
%   %NTODO: Note define D
%   \State Draw a sample $\hat x_s$, from $P_s(x;\mu_s, \Sigma_s)$
%   \State $\hat y_s = \operatorname*{argmax}_c P_c(\hat x_s)  $ 
%   \State $\hat X_s \gets \hat X_s\, ||\, \hat x_s$ 
%   \If{$P_{\hat y_s}(\hat x_s) < P_{\hat y_s}(\mu_{\hat y_s} + 3 *\sqrt{\lambda_{y_s}}*v_{y_s})$}
%         \State $\hat Y_s \gets \hat Y_s\, ||\, \hat y_s$ 
%   \Else    
%         \State $\hat Y_s \gets \hat Y_s\, ||\, |K_s| + 1$ 
% \end{algorithmic}
% \end{algorithm}

%%%%%%%%%%%%%%%%%%%%%%%%%%%%%%%%%%%%%%%%%%%%%%%%%%%%%%%%%%%%%%%%%%%%%%%%%%%%%%%%%%%%%%%%%%%%%%%%%%%%%%%

%%%%%%%%%%%%%%%%%%%%%%%%%%%%%%%%%%%%%%%%%%%%%%%%%%
%                                         _     
%                                        | |    
%   __ _ _ __  _ __  _ __ ___   __ _  ___| |__  
%  / _` | '_ \| '_ \| '__/ _ \ / _` |/ __| '_ \ 
% | (_| | |_) | |_) | | | (_) | (_| | (__| | | |
%  \__,_| .__/| .__/|_|  \___/ \__,_|\___|_| |_|
%       | |   | |                               
%       |_|   |_|  
%%%%%%%%%%%%%%%%%%%%%%%%%%%%%%%%%%%%%%%%%%%%%%%%%%

%%%%%%%%%%%%%% Figure GANTREE %%%%%%%%%%%%%%%%%%%%%%%%%%%%
\begin{figure}[t!]
%\vskip 0.2in
\begin{center}
\centerline{\includegraphics[width=1.0\columnwidth]{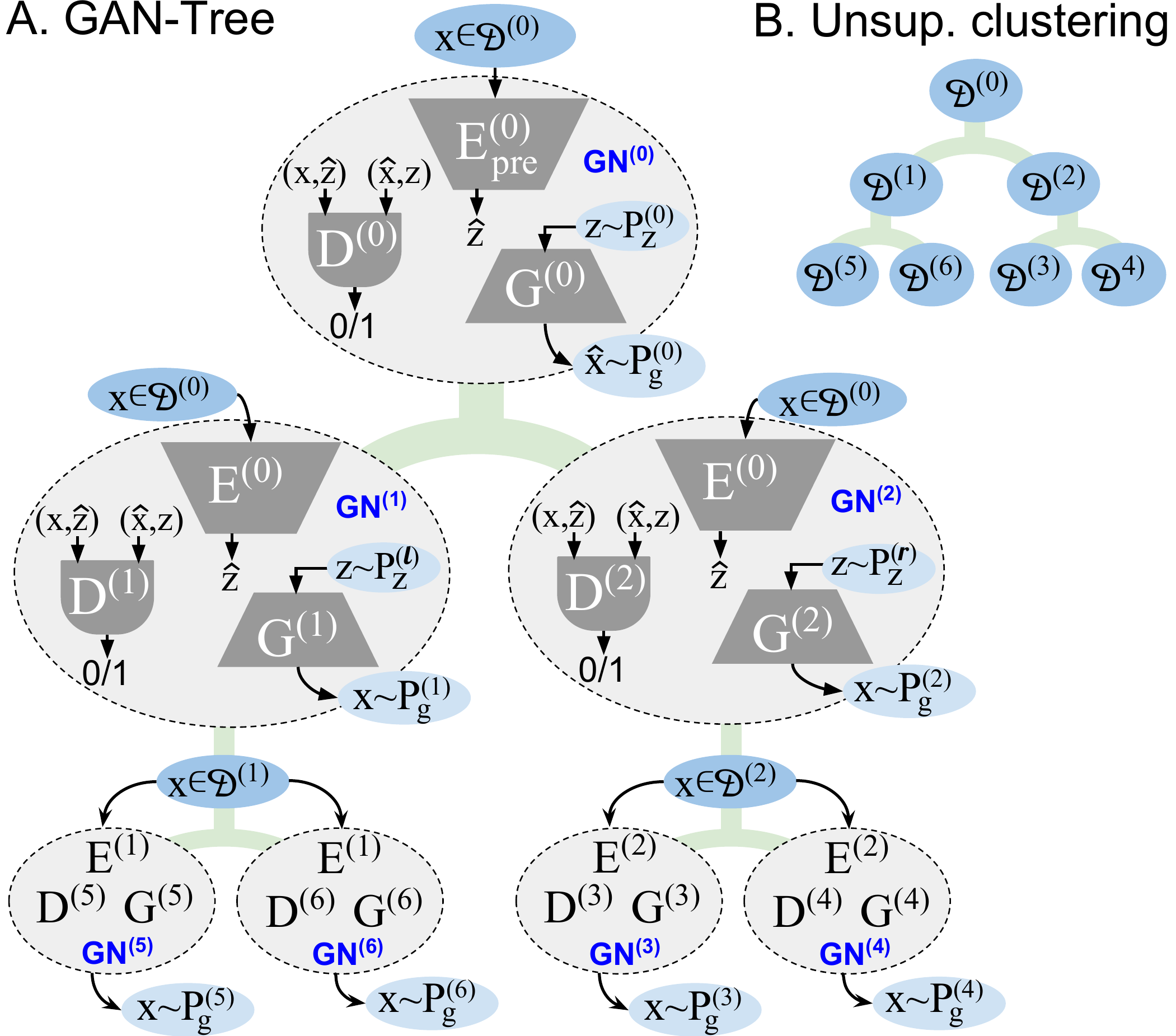}}
%\vspace{-3mm}
\caption{
Illustration of the hierarchical structure of \gantree (part A) and unsupervised clustering of data samples $\mathcal{D}$ in a hierarchical fashion (part B). Composition of a single \gnode at the root level shows how the networks are used in an ALI~\cite{dumoulin2016adversarially} framework inside a \textit{GNode}. %It also illustrates the use of the common encoder \encoder{0}{} by child nodes \textit{GN}$^{(1)}$ and \textit{GN}$^{(2)}$, while utilizing separate copies of generator (\emph{G}) and discriminator (\emph{D}) networks.
}
\label{fig:gantree_concept}
\end{center}
\vspace{-10mm}
\end{figure}
%%%%%%%%%%%%%% Figure GANTREE %%%%%%%%%%%%%%%%%%%%%%%%%%%%

\section{Approach}
\label{section:approach}
In this section, we provide a detailed outline of the construction scheme and training algorithm of \gantree (Section \ref{sec:31}-\ref{sec:33}). Further, we discuss the inference methods for fetching a \ganset from a trained \gantree for generation (Section \ref{sec:34}). We also elaborate on the procedure to incrementally extend a previously trained \gantree using new data samples from a different category (Section \ref{sec:35}).

%In this section, we first provide an outline of the construction scheme and training algorithm of the \gantree framework (Section \ref{sec:31} - \ref{sec:33}). Further, we discuss the inference/query methods for fetching a \ganset from a trained \gantree, utilizing it for generation and clustering tasks (Section \ref{sec:34}). We also elaborate on the procedure to incrementally extend a previously trained \gantree using new data samples from a different category (Section \ref{sec:35}). 

\subsection{Formalization of \textbf{\textit{GAN-Tree}}}
\label{sec:31}

A \gantree is a full binary tree where each node indexed with $i$, \textit{GN$^{(i)}$} (GNode), represents an individual \textit{GAN} framework. The root node is represented as \textit{GN}$^{(0)}$ with the corresponding children nodes as \textit{GN}$^{(1)}$ and \textit{GN}$^{(2)}$ (see Fig.~\ref{fig:gantree_concept}). Here we give a brief overview of a general \gantree framework. Given a set of target samples $\mathcal{D}$ = $(x_i)^n_{i=1}$ drawn from a true data distribution $\mathcal{P}_d$, the objective is to optimize the parameters of the mapping $G:\mathcal{Z} \rightarrow \mathcal{X}$, such that the distribution of generated samples $G(z) \sim \mathcal{P}_g$ approximates the target distribution $\mathcal{P}_d$ upon randomly drawn latent vectors $z \sim \mathcal{P}_z$. Recent generative approaches~\cite{che2016mode} propose to simultaneously train an inference mapping, $E:\mathcal{X} \rightarrow \mathcal{Z}$ to avoid mode-collapse. In this paper, we have used Adversarially Learned Inference (ALI)~\cite{dumoulin2016adversarially} framework as the basic GAN formulation for each node of \gantree. However, one can employ any other GAN framework for training the individual \gantree nodes, if it satisfies the specific requirement of having an inference mapping.

%%%%%%%%%%%%%%%%%%%%% Algorithm 1: GAN_Tree Training Algorithm %%%%%%%%%%%%%%%%%%%%%
{
\begin{algorithm}[!t]
%\vspace{-6mm}
%\algsetup{linenosize=\tiny}
\small %\small, \footnotesize, \scriptsize, or \tiny
\caption{\gantree Construction/Training Algorithm}
\label{algo:gantree}
\begin{algorithmic}[1]
\State \textbf{input:} \gantree tree
\State node $\gets$ createRoot(tree)
\State Train $E^{(0)}_{pre}$, $G^{(0)}$ and $D^{(0)}$ with GAN Training procedure with Unimodal prior $\mathcal{P}^{(0)}_z$ = $\mathcal{N}(0, \mathcal{I})$
\While{CanFurtherSplit(tree)}
    \State{$\mathrm{S} \gets$ LeafNodes(tree)}
    \State {$i \gets \operatorname*{argmin}_{j \in \mathrm{S}}$ $\dfrac{1}{|\mathcal{D}^{(j)}|} \sum_{x \in \mathcal{D}^{(j)}} {p}_g^{(j)}(x)$}
    \State Initialize \encoder{i}{} with params of \encoder{par(i)}{}
    \State Initialize \decoder{l} and \decoder{r} with params of \decoder{i}
    \State Initialize \disc{l} and \disc{r} with params of \disc{i}
    \State $\mathcal{P}^{(l)}_{z} \gets \mathcal{N}(\dfrac{k\sigma}{2\sqrt{d}} \mathbf{1}, \mathcal{I})$, $\mathcal{P}^{(r)}_{z} \gets \mathcal{N}(-\dfrac{k\sigma}{2\sqrt{d}} \mathbf{1}, \mathcal{I})$
    \State {{ModeSplitProcedure} (\textit{\gnode}$^{(i)}$)}
    \State {$\pi^{(l)} \gets \dfrac{|\mathcal{D}^{(l)}|}{|\mathcal{D}^{(i)}|}$, $\pi^{(r)} \gets \dfrac{|\mathcal{D}^{(r)}|}{|\mathcal{D}^{(i)}|}$}
    \State BiModalGAN-Training({\gnode}$^{(i)}$)
\EndWhile
\end{algorithmic}%\vspace{-4mm}
\end{algorithm}
}
%%%%%%%%%%%%%%%%%%%%%%%%%%%%%%%%%%%%%%%%%

\vspace{1mm}
\noindent
\textbf{Root node ($GN^{(0)}$).}
Assuming $\mathcal{D}^{(0)}$ as the set of complete target samples, the root node $GN^{(0)}$ is first trained using a single-mode latent prior distribution $z\sim \mathcal{P}^{(0)}_z$.  As shown in Fig.~\ref{fig:gantree_concept}; $E^{(0)}_{pre}$, $G^{(0)}$ and $D^{(0)}$ are the encoder, generator and discriminator network respectively for the root node with index-$0$; which are trained to generate samples, $x\sim\mathcal{P}_g^{(0)}$ approximating $\mathcal{P}_d^{(0)}$. Here, $\mathcal{P}_d^{(0)}$ is the true target distribution whose samples are given as $x\in\mathcal{D}^{(0)}$. After obtaining the best approximate $\mathcal{P}_g^{(0)}$, %under the constrained boundary of a) computational potential characterized by the network architectures and b) the limitations of assuming a single-mode latent noise distribution; 
the next objective is to improve the approximation by considering the multi-modal latent distribution in the succeeding hierarchy of \gantree.

\vspace{1mm}
\noindent
\textbf{Children nodes ($GN^{(l)}$ and $GN^{(r)}$).}
Without any choice of the initial number of modes, we plan to split each GNode into two children nodes (see Fig.~\ref{fig:gantree_concept}).  In a general setting, assuming $p$ as the parent node index with the corresponding two children nodes indexed as $l$ and $r$, we define $l=left(p)$, $r=right(p)$, $p=par(l)$ and $p=par(r)$ for simplifying further discussions. Considering the example shown in Fig.~\ref{fig:gantree_concept}, with the parent index $p=0$, the indices of left and right child would be $l=1$ and $r=2$ respectively. A novel binary \textit{Mode-splitting} procedure (Section \ref{sec:32}) is incorporated, which, without using the label information, effectively exploits the most discriminative semantic difference at the latent $\mathcal{Z}$ space to realize a clear binary clustering of the input target samples. We obtain cluster-set $\mathcal{D}^{(l)}$ and $\mathcal{D}^{(r)}$ by applying \textit{Mode-splitting} on the parent-set $\mathcal{D}^{(p)}$ such that $\mathcal{D}^{(p)} = \mathcal{D}^{(l)} \cup \mathcal{D}^{(r)}$. Note that, a single encoder $E^{(p)}$ network is shared by both  the child nodes $GN^{(l)}$ and $GN^{(r)}$ as it is also utilized as a routing network, which can route a given target sample $x$ from the root-node to one of the leaf-nodes by traversing through different levels of the full \gantree.  The bi-modal latent distribution at the output of the common encoder $E^{(p)}$ is defined as $z\sim\mathcal{P}_z^{(l)}$ and $z\sim\mathcal{P}_z^{(r)}$for the left and right child-node respectively. 

After the simultaneous training of $GN^{(l)}$ and $GN^{(r)}$ using a Bi-Modal Generative Adversarial Training (\textit{BiMGAT}) procedure (Section \ref{sec:33}), we obtain an improved approximation ($\mathcal{P}_g^{(p)}$) of the true distribution ($\mathcal{P}_d^{(p)}$) as $\mathcal{P}_g^{(p)} = \pi^{(l)} \mathcal{P}_g^{(l)} + \pi^{(r)}\mathcal{P}_g^{(r)}$. Here, the generated distributions $\mathcal{P}_g^{(l)}$ and $\mathcal{P}_g^{(r)}$ are modelled as $G^{(l)}(z\sim\mathcal{P}_z^{(l)})$ and $G^{(r)}(z\sim\mathcal{P}_z^{(r)})$ respectively (Algo.~\ref{algo:gantree}). Similarly, one can split the tree further to effectively capture the inherent number of modes associated with the true data distribution $\mathcal{P}_d$. %This becomes the primary novelty of our approach. Unlike other works which utilize multi-generator networks or multi-mode prior distribution, \gantree readily allows using a fewer number of modes than that modeled, or even further extend the training of the tree to use higher number of modes, giving extreme flexibility to choose or modify the choice of the number of modes with minimal effort. 
% Considering T as the set of terminal leaf GAN node indices the generated distribution can be expressed as, $\mathcal{P}_g = \sum_{i\in T} \pi^{*(i)}\mathcal{P}_g^{(i)}$, where $\pi^{*(0)} = 1$ and $\pi^{*(i)} = \pi^{*(par(i))} * \pi^{(i)}$.

\vspace{1mm}
\noindent
\textbf{Node Selection for split and stopping-criteria.}
A natural question then arises of how to decide which node to split first out of all the leaf nodes present at a certain state of \gantree ? For making this decision, we choose the leaf node which gives minimum mean likelihood over the data samples labeled for it (lines 5-6, Algo.~\ref{algo:gantree}). Also, the stopping criteria on the splitting of \gantree has to be defined carefully to avoid overfitting to the given target data samples. For this, we make use of a robust {IRC-based stopping criteria} \cite{han2007robust} over the embedding space $\mathcal{Z}$, preferred against standard AIC and BIC metrics. However, one may use a fixed number of modes as a stopping criteria and extend the training from that point as and when required.

%                      _                    _ _ _   
%                     | |                  | (_) |  
%  _ __ ___   ___   __| | ___     ___ _ __ | |_| |_ 
% | '_ ` _ \ / _ \ / _` |/ _ \   / __| '_ \| | | __|
% | | | | | | (_) | (_| |  __/   \__ \ |_) | | | |_ 
% |_| |_| |_|\___/ \__,_|\___|   |___/ .__/|_|_|\__|
%                                    | |            
%                                    |_|            
%%%%%%%%%%%%%%%%%%%%%%%%%%%%% Algorithm 2: Mode Split Procedure %%%%%%%%%%%%%%%%%%%%%%%%%%%%%%%%%%%%%%%
\begin{algorithm}[t]
% \fontsize{7pt}{6pt}
% \selectfont
\small
\caption{Mode Split Procedure}
\label{algo:mode_split}
\begin{algorithmic}[1]
\State \textbf{input:} \gnode with index $i$, left and right child $l$ and $r$
\State Initialize unassigned bag $\mathcal{B}_u$ with $\mathcal{D}^{(i)}$, assigned bag $\mathcal{B}_a$ with $\phi$, and cluster label map $\mathrm{L}$ with $\phi$
%\State Initialize assigned bag $\mathcal{B}_a$ with $\phi$
%\State Initialize cluster label map $\mathrm{L}$ with $\phi$
%\Repeat \:$\mathcal{B}_u$ is empty
\While{  $\vert \mathcal{B}_u\vert \neq 0$}
    \For {$n_0$ iterations}
        \State Sample minibatch $x_u$ of $m$ data samples \Statex \hspace{1.0cm} \{$x^{(1)}_u,x^{(2)}_u,...,x^{(m)}_u$\} from $\mathcal{B}_u$.
        \State Sample minibatch $x_a$ of $m$ data samples \Statex \hspace{1.0cm} \{$x^{(1)}_a,x^{(2)}_a,...,x^{(m)}_a$\} from $\mathcal{B}_a$.
        \For{$j$ from $1$ to $m$}
            \State $z^{(j)}_a \gets E^{(i)}(x^{(j)}_a)$; \; $z^{(j)}_u \gets E^{(i)}(x^{(j)}_u)$
            \State ${c}_j = \mathrm{L}(x^{(j)}_a)$ (assigned cluster label)
            \State $t_j = \operatorname*{argmax}_{k \in \{l, r\}} p(k\vert z^{(j)}_u)$ (temp. label)
        \EndFor
        \State $\mathcal{L}_{recon}^{(a)} \gets  \frac{1}{m} \sum_{j=1}^{m} \norm{x^{(j)}_a - G^{(c_j)}(z^{(j)}_a)}^{2}_{2}$
        \State $\mathcal{L}_{recon}^{(u)} \gets  \frac{1}{m} \sum_{j=1}^{m} \norm{x^{(j)}_u - G^{(t_j)}(z^{(j)}_u)}^{2}_{2}$
        \State $\mathcal{L}_{nll}^{(a)} \gets  \frac{1}{m} \sum_{j=1}^{m} -log(p^{(c_j)}_z(z^{(j)}_a))$
        \State $\mathcal{L}_{nll}^{(u)} \gets  \frac{1}{m} \sum_{j=1}^{m} -log(p^{(t_j)}_z(z^{(j)}_u))$
        \State $\mathcal{L}_{split} \gets \mathcal{L}_{recon}^{(a)} + \mathcal{L}_{recon}^{(u)}+ \mathcal{L}_{nll}^{(a)}+ \mathcal{L}_{nll}^{(u)}$
        \State{Update parameters $\Theta_{E^{(i)}}$, $\Theta_{G^{(l)}}$, $\Theta_{G^{(r)}}$ \Statex \hspace{1.0cm} by optimizing $\mathcal{L}_{split}$ using Adam}
    \EndFor
    \For{$x \in \mathcal{B}_u$}
        \If{$\max_{k \in \{l, r\}} p^{(k)}_z(E^{(i)}(x)) > \gamma_0$}
            \State Move $x$ from $\mathcal{B}_u$ to $\mathcal{B}_a$
            \State $\mathrm{L}(x) \gets \operatorname*{argmax}_{k \in \{l, r\}} p(k \vert E^{(i)}(x))$
        \EndIf
    \EndFor
\EndWhile
%\Until{\textup{convergence}}
\end{algorithmic}
\end{algorithm}
%%%%%%%%%%%%%%%%%%%%%%%%%%%%%%%%%%%%%%%%%%%%%%%%%%%%%%%%%%%%%%%%%%%%%%%%%%%%%%%%%%%%%%%%%%%%%%%%%%%%%%%

\subsection{\textbf{\textit{Mode-Split}} procedure}
\label{sec:32}
The \textit{mode-split} algorithm is treated as a basis of the top-down divisive clustering idea, which is incorporated to construct the hierarchical \gantree by performing binary split of individual \gantree nodes. The splitting algorithm must be efficient enough to successfully exploit the highly discriminative semantic characteristics in a fully-unsupervised manner. To realize this, we first define $P_z^{(l)} = \mathcal{N}(\mu^{(l)},\Sigma^{(l)})$ and $P_z^{(r)}=\mathcal{N}(\mu^{(r)},\Sigma^{(r)})$ as the fixed normal prior distributions (non-trainable) for the left and right children respectively. A clear separation between these two priors is achieved by setting the distance between the mean vectors as $k\sigma$  with $\Sigma^{(l)} = \Sigma^{(r)} = \sigma^2\mathcal{I}_d$; where $\mathcal{I}_d$ is a $d\times d$ identity matrix. Assuming $i$ as the parent node index, $\mathcal{D}^{(i)}$ is the cluster of target samples modeled by $GN^{(i)}$. Put differently, the objective of the mode-split algorithm is to form two mutually exclusive and exhaustive target data clusters $\mathcal{D}^{(l)}$ and $\mathcal{D}^{(r)}$, by utilizing the likelihood of the latent representations to the predefined priors $P_z^{(l)}$ and $P_z^{(r)}$. 

%The algorithm involves simultaneous update of three different network parameters viz, $\Theta_{E^{(i)}}$, $\Theta_{G^{(l)}}$, $\Theta_{G^{(r)}}$ as described in Algorithm \ref{algo:mode_split}.

%The algorithm is developed keeping in mind two important aspects. Firstly, the semantic regularity of the bi-modal latent space has to be kept intact during the splitting procedure as it helps to avoid mode-collapse. Secondly, the algorithm must support aggressive splitting of the input target samples by effectively exploiting the most discriminative semantic characteristics. 
To effectively realize \textit{mode-splitting} (Algo.~\ref{algo:mode_split}), we define two different bags; a) assigned bag $\mathcal{B}_a$ and b) unassigned bag $\mathcal{B}_u$. $\mathcal{B}_a$ holds the semantic characteristics of individual modes in the form of representative high confidence labeled target samples. Here, the assigned labeled samples are a subset of the parent target samples, $x\in\mathcal{D}^{(i)}$, with the corresponding hard assigned cluster-id obtained using the likelihood to the predefined priors (line 11, Algo.~\ref{algo:mode_split}) in the transformed encoded space. We refer it as a hard-assignment as we do not update the cluster label of these samples once they are moved from $\mathcal{B}_u$ to the assigned bag, $\mathcal{B}_a$. This effectively tackles mode-collapse in the later iterations of the mode-spilt procedure. For the samples in $\mathcal{B}_u$, a temporary cluster label is assigned depending on the prior with maximum likelihood (line 12, Algo.~\ref{algo:mode_split}) to aggressively move them towards one of the binary modes (lines 19-22, Algo.~\ref{algo:mode_split}). Finally, the algorithm converges when all the samples in $\mathcal{B}_u$ are moved to $\mathcal{B}_a$.
The algorithm involves simultaneous update of three different network parameters (line 18, Algo.~\ref{algo:mode_split}) using a final loss function $\mathcal{L}_{split}$ consisting of:
\begin{itemize}
\item the likelihood maximization term $\mathcal{L}_{nll}$ for samples in both $\mathcal{B}_a$ and $\mathcal{B}_u$ (lines 15-16) encouraging exploitation of a binary discriminative semantic characteristic, and  
\item the semantic preserving reconstruction loss $\mathcal{L}_{recon}$ computed using the corresponding generator i.e. $G^{(l)}$ and $G^{(r)}$ (lines 13-14). This is used as a regularization to hold the semantic uniqueness of the individual samples avoiding \textit{mode-collapse}.
\end{itemize}

\subsection{BiModal Generative Adversarial Training}
\label{sec:33}
The mode-split algorithm does not ensure matching of the generated distribution $G^{(l)}(z\sim\mathcal{P}_z^{(l)})$ and $G^{(r)}(z\sim\mathcal{P}_z^{(r)})$ with the expected target distribution $\mathcal{P}_d^{(l)}$ and $\mathcal{P}_d^{(r)}$ without explicit attention. Therefore to enable generation of plausible samples from the randomly drawn prior latent vectors, a generative adversarial framework is incorporated simultaneously for both left and right children. In ALI~\cite{dumoulin2016adversarially} setting, the loss function involves optimization of the common encoder along with both the generators in an adversarial fashion; utilizing two separate discriminators, which are trained to distinguish $E^{(p)}(x\in\mathcal{D}^{(l)})$ and $E^{(p)}(x\in\mathcal{D}^{(r)})$ from $z\sim \mathcal{P}_z^{(l)}$ and $z\sim\mathcal{P}_z^{(r)}$ respectively. %Algorithm \ref{algo:gantree} demonstrates an overview of the full \gantree training algorithm with sequential call of \textit{mode-split} and \textit{BiMGAT} procedure for each \gantree node.

\subsection{GAN-Set: Generation and Inference}
\label{sec:34}
%*%A fully trained \gantree enables us to select an exponentially large number of \textit{GAN-Set}s, giving flexibility to choose the extent of quality and diversity as required for the subsequent task. 
To utilize a generative model spanning the entire data distribution $\mathcal{P}_d$, an end-user can select any combination of nodes from a fully trained \gantree (i.e. \textit{GAN-Set}) such that the data distribution they model is exhaustive and mutually exclusive. However, to generate only a subset of the full data distribution, one may choose a mutually exclusive, but non-exhaustive set - \textit{Partial} \ganset.

%While choosing a \ganset to perform generation, one must keep in mind the requirement in terms of the quality of generated samples and diversity across various classes in the distribution. 
For a use case where extreme preference is given to diversity in terms of the number of novel samples over quality of the generated samples, selecting a singleton set - \{\textit{root}\} would be an apt choice. However, in a contrasting use case, one may select all the leaf nodes as a \textit{Terminal} \ganset to have the best quality in the generated samples, albeit losing the novelty in generated samples. The most practical tasks will involve use cases where a \ganset is constructed as a combination of both intermediate nodes and leaf nodes.

A \ganset can also be used to perform clustering and label assignment for new data samples in a fully unsupervised setting. We provide a formal procedure \textit{AssignLabel} in the supplementary document for performing the clustering of the data samples using a \gantree.
% For carrying out node label assignment of new samples from the same distribution, for a given \ganset, we perform the following steps: Starting from the root node, we first check if the node exists in the \ganset. If it does, we assign the index of the node as the label to the sample, else we choose the child node of the current node whose prior distribution better models the given sample, and recurse these steps over the selected child node. 

\vspace{1mm}
\noindent
\textbf{How does \gantree differ from previous works?}

\noindent
\textbf{AdaGAN} -
%Inspired from boosting algorithms, AdaGAN~\cite{tolstikhin2017adagan} proposes an iterative procedure, which incrementally address uncovered data modes by introducing new component GANs using sample reweighting technique. 
Sequential learning approach adopted by~\cite{tolstikhin2017adagan} requires a fully-trained model on the previously addressed mode before addressing the subsequent undiscovered samples. %Such algorithm also restricts training of previously addressed mode once it is already discovered. 
As it does not enforce any constraints on the amount of data to be modeled by a single generator network, it mostly converges to more number of modes than that actually present in the data. In contrast, \gantree models a mutually exclusive and exhaustive set at each splitting of a parent node by simultaneously training child generator networks. Another major disadvantage of \textit{AdaGAN} is that it highly focuses on quality rather than diversity (caused by the over-mode division), which inevitably restricts the latent space interpolation ability of the final generative model. %As already discussed, we consider the choice of number of modes to be highly subjective and the algorithm must offer enough flexibility on this aspect depending the requirement in subsequent task.

\noindent
\textbf{DMGAN} -
Khayatkhoei~\etal\cite{khayatkhoei2018disconnected} proposed a disconnected manifold learning generative model using a multi-generator approach. They proposed to start with an overestimate of the initial number of mode components, $n_g$, than the actual number of modes in the data distribution $n_r$. %To effectively learn the required number of modes in the final model they proposed to learn a distribution over the mode prior probabilities which is later optimized to gradually vanish the contribution of over estimated generator networks. %In contrast to \gantree , they follow a bottom-up strategy by initializing the model from over estimated mode components. 
As discussed before, we do not consider the existence of a definite value for the number of actual modes $n_r$ as considered by \textit{DMGAN}, especially for diverse natural image datasets like CIFAR and ImageNet. In a practical scenario, one can not decide the initial value of $n_g$ without any clue on the number of classes present in the dataset. %Also, the number of classes is also highly subjective as one can consider all animals as one category whereas each animals can also be categorized separately. 
\textit{DMGAN} will fail for cases where $n_g < n_r$ as discussed by the authors. Also note that unlike \gantree, \textit{DMGAN} is not suitable for incremental future expansion. This clearly demonstrates the superior flexibility of \gantree against \textit{DMGAN} as a result of the adopted top-down divisive strategy.
%as \gantree follows a top-down approach, devoid of such shortcoming.

%%%%%%%%%%%%%%%%%%%%%%%%%% Algorithm 3: Incremental Node Training Procedure %%%%%%%%%%%%%%%%%%%%%%%%%%%%%%%%
\begin{algorithm}[!b]
% \fontsize{7pt}{6pt}
% \selectfont
\small
\caption{Incremental Node Training}
\label{algo:ignode}
\begin{algorithmic}[1]
\State \textbf{input:} Node Index $c$, New Data Sample set $\mathcal{D}'$
\State gset = CreateTerminalGanSet(\gnode$^{par(c)}$)
\State Populate an empty bag $\mathcal{B}_a$ of assigned samples with all samples from $\mathcal{D}'$
\State Generate $|\mathcal{D}'|\cdot|$gset$|$ samples from $\mathcal{P}^{(gset)}_g$ and add them to $\mathcal{B}_a$; assign cluster labels based on the corresponding ancestor among the child nodes of \gnode$^{par(c)}$ to $\mathrm{L}$
\State Run \textit{Mode Split Procedure} on \gnode$^{(par(c))}$ training only $E^{(par(c))}$ and $G^{(c)}$ over samples from $\mathcal{B}_a$
\State Run \textit{BiModalGAN-Training} over \textit{GN}$^{(par(c))}$ training only $E^{(par(c))}$, $G^{(c)}$ and $D^{(c)}$
\State Re-evaluate $\pi^{(left(par(c)))}$ and $\pi^{(right(par(c)))}$ 
\hspace{-200mm}
\end{algorithmic}
\end{algorithm}
%%%%%%%%%%%%%%%%%%%%%%%%%%%%%%%%%%%%%%%%%%%%%%%%%%%%%%%%%%%%%%%%%%%%%%%%%%%%%%%%%%%%%%%%%%%%%%%%%%%%%%%

% \begin{figure*}[t]%[h!]
% \centering    
% 	\includegraphics[width=0.99\linewidth]{figures/cfig_3_toy.pdf}
% 	\vspace{-10pt}
% 	\caption{%\fontsize{7}{1}\selectfont 
% 	\textbf{Part A}: Illustration of an ideal transformation function (in green) - a non-invertible mapping from a disconnected uniform distribution to a uni-modal Gaussian distribution and its invertible approximation (in dotted) learned by a neural network. Using the learned approximate function for transforming real data distribution (left) to its latent representation introduces a discontinuity in latent space distribution (top), whose inverse when used for generating real data samples from a prior uni-modal distribution (bottom) reveals generation of bad samples indicated in purple box. \textbf{Part B}: Illustration of the entire \gantree training algorithm procedure over a toy dataset. Initially the entire data distribution (in green box) is modeled by a uni-modal latent distribution; generated distribution (in red) is shown below that. \textbf{Part C}: Generations produced by the root node and its children of a GAN-tree trained on FaceBed dataset, as mention in Section \ref{section:experiments}.
% 	} 
% 	\vspace{-10pt}
% 	\label{fig:toy_gantree}    
% \end{figure*}

\subsection{Incremental \textbf{\gantree}: \textbf{\textit{iGANTree}}}
\label{sec:35}

We advance the idea of \gantree to \igantree, wherein we propose a novel mechanism to extend an already trained \gantree $\mathcal{T}$ to also model samples from a set $\mathcal{D}'$ of new data samples. An outline of the entire procedure is provided across Algorithms \ref{algo:ignode} and \ref{algo:igantree}. To understand the mechanism, we start with the following assumptions from the algorithm. On termination of this procedure over $\mathcal{T}$, we expect to have a single leaf node which solely models the distribution of samples from $\mathcal{D}'$; and other intermediate nodes which are the ancestors of this new node, should model a mixture distribution which also includes samples from $\mathcal{D}'$.

%%%%%%%%%%%%%%%%%%%%%%%%%% Algorithm 4: Incremental Gan Tree Training Procedure %%%%%%%%%%%%%%%%%%%%%%%%%%%%%%%%
% \vspace{-7mm}
\begin{algorithm}[!t]
% \caption{Incremental Node Training}
% \label{algo:igantree}
\small
\caption{Incremental \gantree Training}
\label{algo:igantree}
\begin{algorithmic}[1]
\State \textbf{input:} \gantree $T$, New Data Sample set $\mathcal{D}'$
\State $i \gets$ index(root($T$))
\While{$i$ is NOT a leaf node}
    \State $l \gets$ left($i$); $r \gets$ right($i$)
    \If {Avg($p_x^{(l)}$($\mathcal{D}'$)) $\geq p_z^{(l)}({\mu}^{(l)} + d_{\sigma_0})$}
        %\State 
        $i \gets l$
    \ElsIf{Avg($p_x^{(r)}$($\mathcal{D}'$)) $\geq p_z^{(r)}({\mu}^{(r)} + d_{\sigma_0})$}
        %\State 
        $i \gets r$
    \Else
        %\State 
        \hspace{1mm} break
    \EndIf
    \vspace{1mm}
\EndWhile 
\State Here, $i$ is the current node index
\State $j \gets$ NewId() (new parent index)
\State $k \gets$ NewId() (new child index)
\State par($k$) $\gets j$; \hspace{1mm}par($j$) $\gets$ par($i$); \hspace{1mm}par($i$) $\gets j$
%\State par($j$) $\gets$ par($i$) 
%\State par($i$) $\gets j$
\If {$i$ = index(root($T$))}
    \State root($T$) $\gets j$; \hspace{1mm}left($j$) $\gets i$; \hspace{1mm}right($j$) $\gets k$
    %\State left($j$) $\gets i$
    %\State right($j$) $\gets k$
\ElsIf{$i$ was the left child of its previous parent}
    \State left(par($j$)) $\gets j$; \hspace{1mm}left($j$) $\gets i$; \hspace{1mm}right($j$) $\gets k$
    %\State left($j$) $\gets i$
    %\State right($j$) $\gets k$
\Else
    \State right(par($j$)) $\gets j$; \hspace{1mm} left($j$) $\gets k$; \hspace{1mm} right($j$) $\gets i$
    %\State left($j$) $\gets k$
    %\State right($j$) $\gets i$
\EndIf

\State Create networks $G^{(k)}$, $D^{(k)}$ with random initialization
\State Train $E^{(j)}, G^{(k)}$ and $D^{(k)}$ with $\mathcal{L}_{recon}$ and $\mathcal{L}_{adv}$

\State $E^{(par(j))} \gets$ copy($E^{(par(i))}$)
\State $G^{(j)} \gets$ copy($G^{(i)}$); $D^{(j)} \gets$ copy($D^{(i)}$); $i \gets$ par($i$) 
    
\While{\textit{GN}$^{(i)}$ is not root($T$)}
    \State IncrementalNodeTrain($i$); $i \gets$ par($i$)
\EndWhile
\textbf{end while}
\State Train \textit{GN}$^{(i)}$ with GAN Training Procedure on $\mathcal{D}$' and generated samples from Terminal GAN-Set.

\end{algorithmic}
\end{algorithm}
%%%%%%%%%%%%%%%%%%%%%%%%%%%%%%%%%%%%%%%%%%%%%%%%%%%%%%%%%%%%%%%%%%%%%%%%%%%%%%%%%%%%%%%%%%%%%%%%%%%%%%%

To achieve this, we first find out the right level of hierarchy and position to insert this new leaf node using a seek procedure (lines 2-8 in Algo.~\ref{algo:igantree}). Here $p_x^{(l)}(x) = p_z^{(l)}(E^{(l)}(x))$ and similarly for $r$, in lines 5-6. Let's say the seek procedure stops at node index $i$. We now introduce 2 new nodes \textit{GN}$^{(j)}$ and \textit{GN}$^{(k)}$ in the tree and perform reassignment (lines 11-17 in Algo.~\ref{algo:igantree}). The new child node \textit{GN}$^{(k)}$ models only the new data samples; and the new parent node \textit{GN}$^{(j)}$ models a mixture of $\mathcal{P}^{(i)}_g$ and $\mathcal{P}^{(k)}_g$. %It becomes clear from this, that \textit{GN}$^{(j)}$ needs to be the child of \textit{GN}$^{(par(i))}$; and the parent of \textit{GN}$^{(i)}$ and \textit{GN}$^{(k)}$. We now perform a reassignment (lines 13-22 in Algo~\ref{algo:igantree}).
This brings us to the question, how do we learn the new distribution modeled by \textit{GN}$^{(par(i))}$ and its ancestors? To solve this, we follow a bottom-up training approach from \textit{GN}$^{(par(i))}$ to \textit{GN}$^{(root(\mathcal{T}))}$, incrementally training each node on the branch with samples from $\mathcal{D}$' to maintain the hierarchical property of the \gantree (lines 22-24, Algo. \ref{algo:igantree}).

Now, the problem reduces down to retraining the parent \encoder{p}{} and the child \decoder{c} and \disc{c} networks at each node in the selected branch, such that (i) \encoder{p}{} correctly routes the generated data samples $x$ to the proper child node and (ii) the samples from $\mathcal{D}'$ are modeled by the new distribution $\mathcal{P}'_g$ at all the ancestor nodes of \textit{GN}$^{(k)}$, remembering the samples from distribution $\mathcal{P}_g$ at the same time. Moreover, we make no assumption of having the data samples $\mathcal{D}$ on which the \gantree was trained previously. To solve the problem of training the node \textit{GN}$^{(i')}$, we make use of terminal \ganset of the sub \gantree rooted at \textit{GN}$^{(i')}$ to generate samples for retraining the node. A thorough procedure of how each node is trained incrementally is illustrated in Algo.~\ref{algo:ignode}. Also, note that we use the mean likelihood measure to decide which of the two child nodes has the potential to model the new samples. We select the child whose mean vector has the minimum average Mahalanobis distance ($d_{\sigma}$) from the embeddings of the samples of $\mathcal{D}$'. This idea can also be implemented to have a full persistency over the structure~\cite{driscoll1989making} (further details in  Supplementary).

% Now, in order to have full persistency of the structure, instead of retraining, we make a copy of the nodes which are to be modified, initialized from their previous version's counterpart, and train them. This way, every incremental addition produces a new root for the new version of \gantree with the same time complexity, and additional space complexity equivalent to the space occupied by the nodes in a single branch, while maintaining full persistency of the previously trained versions of the \gantree.

%%%%%%%%%%%%%%%%%%%%%%%% figure igantree %%%%%%%%%%%%%%%%%%%%%%
\begin{figure}[!t]
%\vskip 0.2in
\begin{center}
\centerline{\includegraphics[width=0.8\columnwidth]{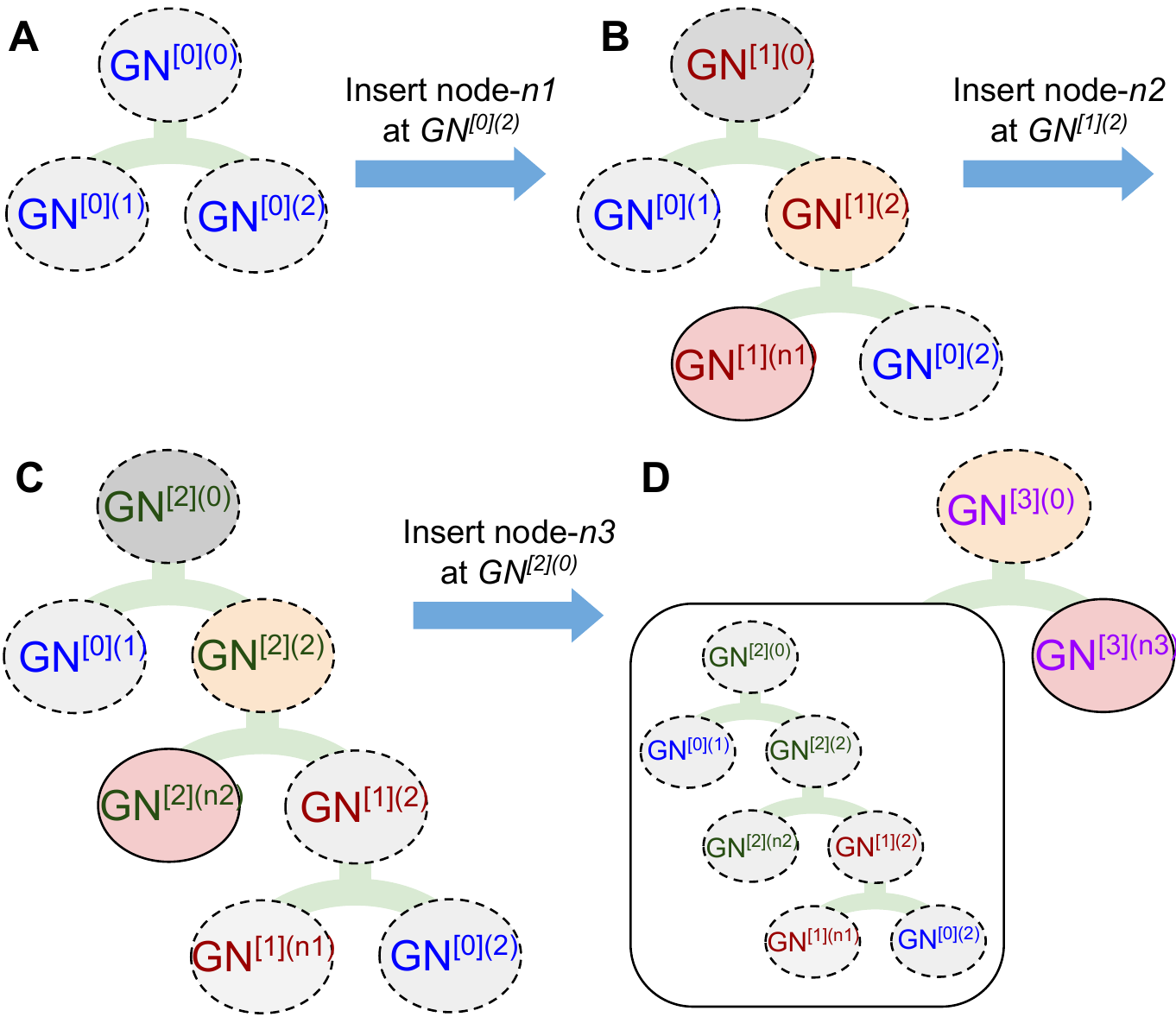}}
\caption{%\fontsize{7}{1}\selectfont %Snapshots of the different versions of \gantree after running Incremental GanTree Training (Algorithm \ref{algo:igantree}). 
Snapshots of the different versions of incrementally obtained \gantree. Here A is the pretrained \gantree over which Algo.~\ref{algo:igantree} is run to obtain B, and subsequently C and D. Each transition highlights the branch which is updated in gray, with the new child node in red, the new parent node in orange, while the rest of the nodes stay intact. In B, nodes labeled with red are the ones which are updated. Similarly, in C and D, the updated nodes are labeled with green and purple respectively. It is illustrated that just by incrementally adding a new branch by updating nodes from its previous version, it exploits the full persistence of the \gantree and provides all the versions of root nodes - \textit{GN}$^{[0:4](0)}$. \vspace{-5mm}}
\label{icml-historical}
\end{center}
\vskip -0.2in
\end{figure}
%%%%%%%%%%%%%%%%%%%%%%%%  %%%%%%%%%%%%%%%%%%%%%%

%%%%%%%%%%%%%%%%%%%%%%%% Figure 2 %%%%%%%%%%%%%%%%%%%%%
\begin{figure*}[t]%[h!]
\centering    
	\includegraphics[width=1.0\linewidth]{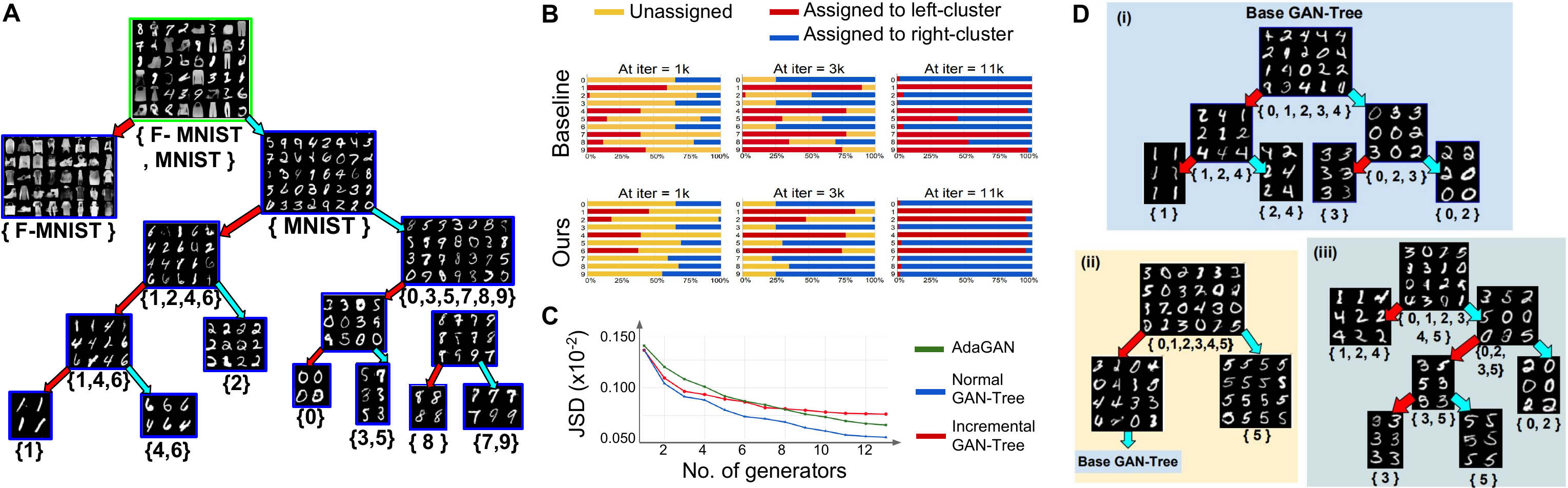}%cfig_4_final_copy
	\vspace{-5pt}
	\caption{%\fontsize{7}{1}\selectfont 
	\textbf{Part A}: Illustration of the \gantree training procedure over MNIST+Fashion-MNIST dataset. \textbf{Part B}: Effectiveness of our \textit{mode-split} procedure (with bagging) against the baseline deep-clustering technique (without bagging) on MNIST root node. Our approach divides the digits into two groups in a much cleaner way (at iter=11k). \textbf{Part C}: We evaluate the \gantree and \igantree algorithms against the prior incremental training method AdaGAN~\cite{tolstikhin2017adagan}. We train up to 13 generators and evaluate their mean JS Divergence score (taken over 5 repetitions). \textbf{Part D}: Incremental \gantree training procedure \textbf{(i)} Base GAN-Tree, trained over digits 0-4 \textbf{(ii)} GAN-Tree after addition of digit \textbf{5}, with $d_{\sigma_0}$ = 4 \textbf{(iii)} GAN-Tree after addition of digit \textbf{5}, with $d_{\sigma_0}$ = 9.} 
	\label{fig:mnist_incremental}  
	\vspace{-10pt}
\end{figure*}
%%%%%%%%%%%%%%%%%%%%%%%% Figure 2 ends %%%%%%%%%%%%%%%%%

%%%%%%%%%%%%%%%%%%%%%%%% Figure 1b %%%%%%%%%%%%%%%%%%%%%
\begin{figure}[t]%[h!]
\centering    
	\includegraphics[width=0.92\linewidth]{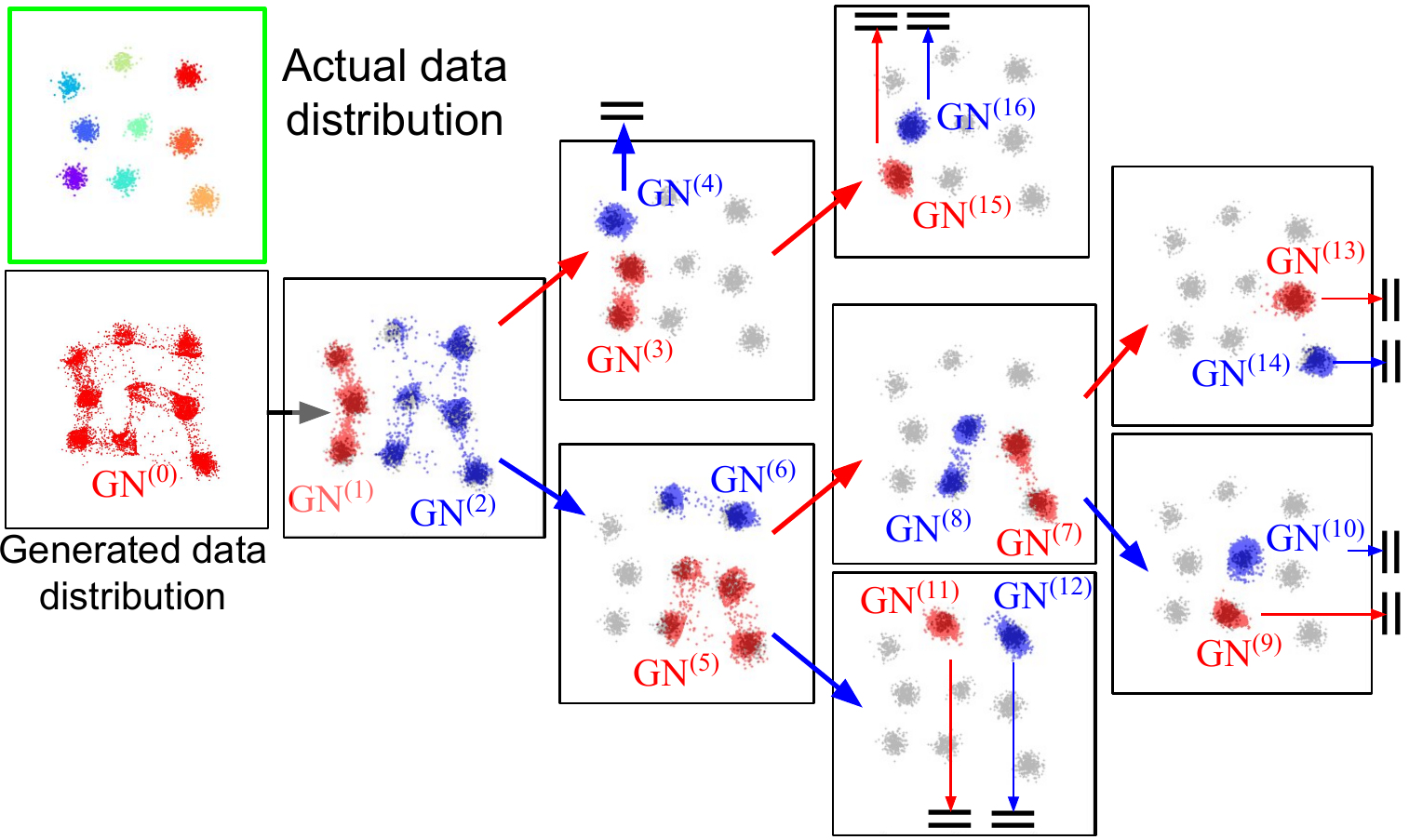}%cfig_4_final_copy
	\vspace{-5pt}
	\caption{%\fontsize{7}{1}\selectfont 
	Illustration of the \gantree progression over the toy dataset. %Initially the entire data distribution (in green box) is modeled by a uni-modal latent distribution; generated distribution (in red) is shown below that.
	} 
	\label{fig:toy}  
	\vspace{-12pt}
\end{figure}
%%%%%%%%%%%%%%%%%%%%%%%% Figure 1b ends %%%%%%%%%%%%%%%%%

\section{Experiments}
\label{section:experiments}
In this section, we discuss a thorough evaluation of \gantree against baselines and prior approaches. We decide not to use any improved learning techniques (as proposed by SNGAN~\cite{miyato2018spectral} and SAGAN~\cite{zhang2018self}) for the proposed \gantree framework to have a fair comparison against the prior art~\cite{khayatkhoei2018disconnected,ghosh2017multi,hoang2018mgan} targeting multi-modal distribution. 

\gantree is a multi-generator framework, which can work on a multitude of basic GAN formalizations (like AAE~\cite{44904}, ALI~\cite{dumoulin2016adversarially}, RFGAN~\cite{bang2018high} etc.) at the individual node level. However, in most of the experiments we use ALI~\cite{dumoulin2016adversarially} except for CIFAR, where both ALI~\cite{dumoulin2016adversarially} and  RFGAN~\cite{bang2018high} are used to demonstrate generalizability of \gantree over varied GAN formalizations. Also note that we freeze parameter update of lower layers of encoder and discriminator; and higher layers of the generator (close to data generation layer) in a systematic fashion, as we go deeper in the \gantree hierarchical separation pipeline. Such a parameter sharing strategy helps us to remove overfitting at an individual node level close to the terminal leaf-nodes.

We employ modifications to the commonly used DCGAN~\cite{radford2015unsupervised} architecture for generator, discriminator and encoder networks while working on image datasets i.e. MNIST (32$\times32$), CIFAR-10 (32$\times32$) and Face-Bed (64$\times64$)). However, unlike in DCGAN, we use batch normalization~\cite{ioffe2015batch} with Leaky ReLU non-linearity inline with the prior multi-generator works~\cite{hoang2018mgan}. While training \gantree on Imagenet~\cite{russakovsky2015imagenet}, we follow the generator architecture used by SNGAN~\cite{miyato2018spectral} for a generation resolution of 128$\times$128 with RFGAN~\cite{bang2018high} formalization. For both \textit{mode-split} and \textit{BiModal-GAN} training we employ Adam optimizer~\cite{kingma2014adam} with a learning rate of 0.001. 

%%%%%%%%%%%%%%%%%%%%%%%% Figure 3 %%%%%%%%%%%%%%%%%%%%%
\begin{figure*}[t]%[h!]
\centering    
	\includegraphics[width=0.92\linewidth]{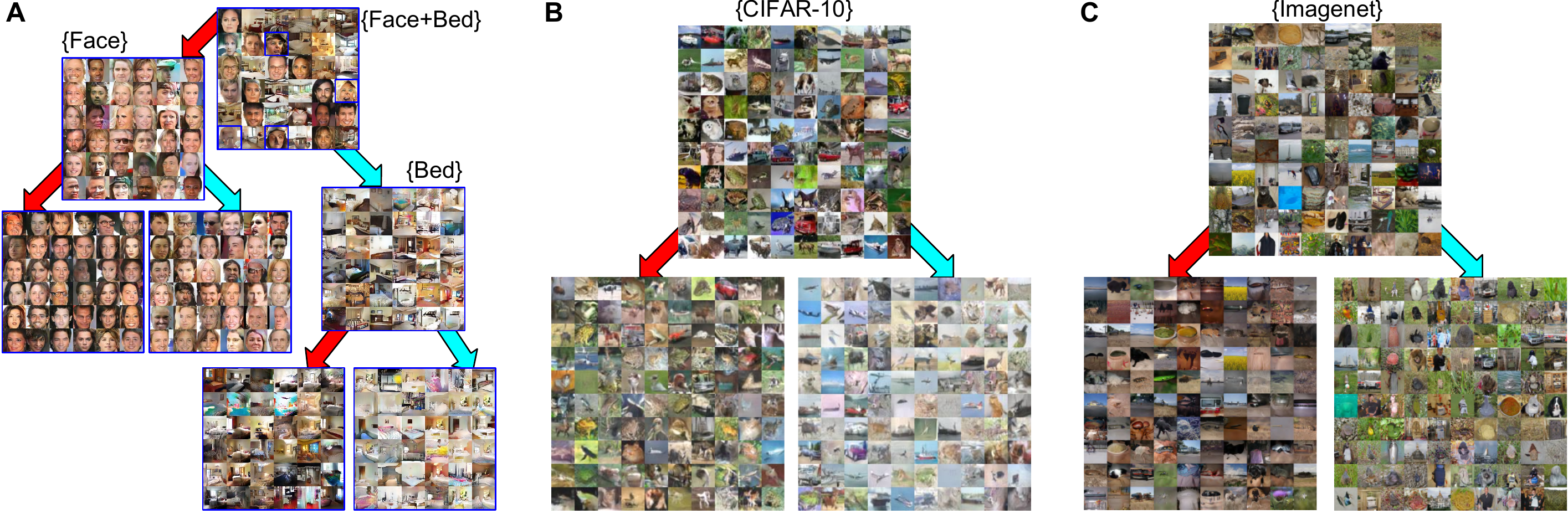}%cfig_4_final_copy
	\vspace{-5pt}
	\caption{%\fontsize{7}{1}\selectfont 
	Generation results on RGB image datasets \textbf{A}: FaceBed,  \textbf{B}:  CIFAR-10, \textbf{C}:  ImageNet. The root node generations of FaceBed show a few implausible generations, which are reduced with further splits. The left child of the root node generates faces, while the right child generates beds. Further splitting the face node, we see that one child node generates images with darker background or darker hair colour, while the other generates images with lighter background or lighter hair colour. Similar trends are observed in the splits of Bed Node in \textbf{Part A}, and also in child nodes of CIFAR-10 and ImageNet. %, where splits are made based on the background variations.
	}
	\label{fig:imagenet}  
	\vspace{-10pt}
\end{figure*}
%%%%%%%%%%%%%%%%%%%%%%%% Figure 3 ends %%%%%%%%%%%%%%%%%

\vspace{1mm}
\noindent
\textbf{Effectiveness of the proposed \textit{mode-split} algorithm.} To verify the effectiveness of the proposed \textit{mode-split} algorithm, we perform an ablation analysis against a baseline deep-clustering~\cite{tian2017deepcluster} technique on the 10-class MNIST dataset. Performance of \gantree highly depends on the initial binary split performed at the root node, as an error in cluster assignment at this stage may lead to multiple-modes for a single image category across both the child tree hierarchy. Fig.~\ref{fig:mnist_incremental}B clearly shows the superiority of \textit{mode-split} procedure when applied at the MNIST root node. 

%%%%%%%%%%%%%%%%%%% Table 3 %%%%%%%%%%%%%%%%%%%%%%
\begin{table}%[!h]
%\vspace{-3mm}
 \caption{Comparison of inter-class variation (JSD) for MNIST ($\times 10^{-2}$) and Face-Bed ($\times 10^{-4}$); and FID score on Face-Bed  inline with~\cite{khayatkhoei2018disconnected}. \vspace{-3mm}
 }
\centering
\setlength\tabcolsep{3pt}
\resizebox{1.01\linewidth}{!} {
\begin{tabular}{l|c|c|c|c}%|ccc
\hline {Model} &\#Gen &JSD MNIST &JSD Face-Bed &FID Face-Bed\\ \hline\hline
%WGAN-GP~\cite{gulrajani2017improved}&1 &$0.14\pm 0.06$ &$0.21\pm 0.15$ &$8.01\pm 0.29$\\
DMWGAN~\cite{khayatkhoei2018disconnected}&20 &$0.21\pm 0.05$ &$0.42\pm 0.23$ &$7.58\pm 0.10$\\
DMWGAN-PL~\cite{khayatkhoei2018disconnected}&20 &$0.08\pm 0.03$ &$0.11\pm 0.06$ &$7.30\pm 0.12$\\
\hline
Ours \ganset &5 &$0.08\pm 0.02$ &$0.10\pm 0.06$ &$7.20\pm 0.11$\\
Ours \ganset &10 &$0.06\pm 0.02$ &$0.09\pm 0.04$ &$7.00\pm 0.10$\\
\hline                        
\end{tabular}
}
\label{table:tab1}
\vspace{-4mm}
\end{table}
%%%%%%%%%%%%%%%%%%%%%%%%%%%%%%%%%%%%%%%%%%%%%%%%%

\vspace{1mm}
\noindent
\textbf{Evaluation on Toy dataset.} We construct a synthetic dataset by sampling 2D points from a mixture of nine disconnected Gaussian distributions with distinct means and covariance parameters. The complete \gantree training procedure over this dataset is illustrated in Fig.~\ref{fig:toy}. As observed, the distribution modeled at each pair of child nodes validates the mutually exclusive and exhaustive nature of child nodes for the corresponding parent.

\vspace{1mm}
\noindent
\textbf{Evaluation on MNIST.} We show an extensive comparison of \gantree against DMWGAN-PL~\cite{khayatkhoei2018disconnected} across various qualitative metrics on MNIST dataset. Table~\ref{table:tab1} shows the quantitative comparison of inter-class variation against previous \textit{state-of-the-art} approaches. It highlights the superiority of the proposed \gantree framework. 

\vspace{1mm}
\noindent
\textbf{Evaluation on compositional-MNIST.} As proposed by Che~\etal~\cite{che2016mode}, the compositional-MNIST dataset  consists of 3 random digits at 3 different quadrants of a full 64$\times$64 resolution template, resulting in a data distribution of 1000 unique modes. Following this, a pre-trained MNIST classifier is used for recognizing digits from the generated samples, to compute the number of modes covered while generating from all of the 1000 variants. Table~\ref{table:tab2} highlights the superiority of \gantree against MAD-GAN~\cite{ghosh2017multi}.

\vspace{1mm}
\noindent
\textbf{\igantree on MNIST.}
We show a qualitative analysis of the generations of a trained \gantree after incrementally adding data samples under different settings. We first train a \gantree for 5 modes on MNIST digits 0-4. We then train it incrementally with samples of the digit 5 and show how the modified structure of the \gantree looks like. Fig.~\ref{fig:mnist_incremental}D shows a detailed illustration for this experiment.

\vspace{1mm}
\noindent
%\textbf{Evaluation on mixed MNIST+F-MNIST and Face-Bed}: 
\textbf{\gantree on MNIST+F-MNIST and Face-Bed.} We perform the divisive \gantree training procedure on two mixed datasets. For MNIST+Fashion-MNIST, we combine 20K images from both the datasets individually. Similarly, following~\cite{khayatkhoei2018disconnected}, we combine Face-Bed to demonstrate the effectiveness of \gantree to model diverse multi-modal data supported on a disconnected manifold (as highlighted by Table~\ref{table:tab1}). The hierarchical generations for MNIST+F-MNIST and the mixed Face-Bed datasets are shown in Fig.~\ref{fig:mnist_incremental}A and Fig.~\ref{fig:imagenet}A respectively.

%%%%%%%%%%%%%   Compositional MNIST   %%%%%%%%%%%%%%%%%%%
\begin{table}%[t]
\caption{Comparison of \gantree against state-of-the-art GAN approaches on compositional-MNIST dataset inline with~\cite{ghosh2017multi}. \vspace{-3mm}}
\centering
\setlength\tabcolsep{4pt}
\resizebox{0.74\linewidth}{!} {
\begin{tabular}{l|c| c}
\hline
{Methods} & {KL Div.$\downarrow$} & {Modes covered $\uparrow$} \\ 
\hline \hline
%DCGAN~\cite{radford2015unsupervised}& 0.18 &980 \\
%BEGAN~\cite{berthelot2017began}& 0.19 &999 \\
WGAN~\cite{arjovsky2017towards}& 0.25 &1000 \\
%Unrolled GAN~\cite{metz2016unrolled} & 0.091 &1000\\
%Info-GAN~\cite{chen2016infogan} &0.47 &990 \\
MAD-GAN~\cite{ghosh2017multi} &0.074 &1000\\ \hline
\ganset (root) &{0.16} &{980} \\
\ganset (5 G-Nodes) &{0.10} &{1000} \\
\ganset (10 G-Nodes) &\textbf{0.072} &\textbf{1000} \\
\hline
\end{tabular}}
\label{table:tab2}
\vspace{-4mm}
\end{table}
%%####################################################

\vspace{1mm}
\noindent
\textbf{On CIFAR-10 and ImageNet.} 
In Table~\ref{table:tab3}, we report the inception score~\cite{salimans2016improved} and FID~\cite{heusel2017gans} obtained by \gantree against prior works on both CIFAR-10 and ImageNet dataset. We separately implement the prior multi-modal approaches, a) GMVAE~\cite{dilokthanakul2016deep} b) ClusterGAN~\cite{mukherjee2019clustergan}, and also the prior multi-generator works,  a) MADGAN~\cite{ghosh2017multi} b) DMWGAN-PL~\cite{khayatkhoei2018disconnected} with a fixed number of generators. Additionally, to demonstrate the generalizability of the proposed framework with varied GAN formalizations at the individual node-level, we implement \gantree with ALI~\cite{dumoulin2016adversarially}, RFGAN~\cite{bang2018high}, and BigGAN~\cite{brock2018large} as the basic GAN setup. Note that, we utilize the design characteristics of BigGAN without accessing the class-label information, along with RFGAN's encoder for both CIFAR-10 and ImageNet. 

%The quantitative comparison in Table~\ref{table:tab3} clearly highlights our superiority even with lesser number of generators (see \textit{\#Gen} column). 

In Table~\ref{table:tab3}, all the approaches targeting ImageNet dataset use modified \textit{ResNet-50} architecture, where the total number of parameter varies depending on the number of generators (considering the hierarchical weight sharing strategy) as reported under the \textit{\#Param} column. While comparing generation performance, one needs access to a selected \ganset instead of the entire \gantree. In Table~\ref{table:tab3}, the performance of \ganset(RFGAN) with 3 generators (i.e. \gantree with total 5 generators) is superior to DMWGAN-PL~\cite{khayatkhoei2018disconnected} and MADGAN~\cite{ghosh2017multi}, each with 10 generators. This clearly shows the superior computational efficiency of \gantree against prior multi-generator works. An exemplar set of generated images with the first root node split is presented in Fig.~\ref{fig:imagenet}B and~\ref{fig:imagenet}C.

\begin{table}[!t]
 \caption{Inception (IS) and FID scores on CIFAR-10 and Imagenet dataset computed on 5K with varied number of generators. %Each experiment in CIFAR is repeated 3 times.
 \vspace{-3mm}}
\centering
\setlength\tabcolsep{2.8pt}
\resizebox{1.01\linewidth}{!}{
\begin{tabular}{l|c|cc|ccc}%|ccc
\hline
\multirow{2}{*}{Method}&\multirow{2}{*}{\#Gen} &\multicolumn{2}{c|}{\tabincell{c}{CIFAR-10}} &\multicolumn{3}{c}{\tabincell{c}{ImageNet}} \\
\cline{3-7} & & IS $\uparrow$ & FID $\downarrow$ &IS $\uparrow$ & FID $\downarrow$ &  \#Param\\\hline\hline
%GMVAE (Dilokthanakul~\etal)& 1& 6.89 & 39.2& -& -& - \\
GMVAE~\cite{dilokthanakul2016deep}& 1& 6.89 & 39.2& -& -& - \\
ClusterGAN~\cite{mukherjee2019clustergan}& 1& 7.02 & 37.1& -& -& - \\
RFGAN~\cite{bang2018high} (\textit{root-node}) &1& 6.87 & 38.0 & 20.01&46.4 & 50M\\
BigGAN (w/o label) & 1& 7.19 & 36.7& 20.89 & 42.5 & 50M  \\
MADGAN~\cite{ghosh2017multi}& 10& 7.33 & 35.1& 20.92& 38.3 & 205M\\
DMWGAN-PL~\cite{khayatkhoei2018disconnected} & 10& 7.41 & 33.1& 21.57& 37.8 & 205M\\\hline

%WGAN-GP~\cite{gulrajani2017improved}&1& 6.53$\pm$.08 & 41.1 & - & -\\
%RFGAN~\cite{bang2018high}(\textit{root-node}) &1& 6.87$\pm$.09 & 38.0 & 20.01&46.4\\
%DMWGAN-PL~\cite{khayatkhoei2018disconnected}&10&7.57$\pm$.11&32.4&-&-\\
%MADGAN~\cite{ghosh2017multi}&10&7.40$\pm$.07& 34.7& -&-\\\hline

Ours \ganset(ALI) &3&7.42&32.5&-&-&-\\
Ours \ganset(ALI) &5&7.63&{28.2}&-&-&-\\
\hline
Ours \ganset(RFGAN) &3&7.60 &28.3& 21.97& 34.0 & 65M\\
Ours \ganset(RFGAN) &5& \textbf{7.91} & \textbf{27.8}& \textbf{24.84}& \textbf{29.4} & \textbf{105M}\\\hline

Ours \ganset(BigGAN) &3& 8.12& 25.2& 22.38& 31.2 & 130M\\
Ours \ganset(BigGAN) &5& \textbf{8.60}& \textbf{21.9}& \textbf{25.93}& \textbf{27.1} & 210M\\

%Ours \ganset(RFGAN) &10&8.02$\pm$.10&\textbf{26.8}&\textbf{23.26}&\textbf{34.8}\\
\hline                        
\end{tabular}
}
\label{table:tab3}
\vspace{-4mm}
\end{table}
%%%%%%%%%%%%%%%%%%%%%%%%%%%%%%%%%%%%%%%%%%%%%%%%%

\section{Conclusion}
\textit{GAN-Tree} is an effective framework to address natural data distribution without any assumption on the inherent number of modes in the given data. Its hierarchical tree structure gives enough flexibility by providing \textit{GAN-Set}s of varied quality-vs-diversity trade-off.
%\textit{GAN-Tree} can also be utilized to acquire inherent semantic clusters from the data samples in a fully unsupervised manner as a result of the mutually exclusive and exhaustive constraint. 
%High degree of flexibility 
This also makes \gantree a suitable candidate for incremental generative modeling. Further investigation on the limitations and advantages of such a framework will be explored in the future.

\vspace{2mm}
\noindent
% \textbf{Acknowledgements.} This work was partially supported by ISRO, Government of India.
\textbf{Acknowledgements.} This work was supported by a Wipro PhD Fellowship (Jogendra) and a grant from ISRO, India.

\newpage
% {\small
% \bibliographystyle{ieee_fullname}
% \bibliography{egbib}
% }

%\includepdf[pages=1-1]{pdf_blank.pdf} 


\section*{Supplementary material}

# GAN-Tree: An Incrementally Learned Hierarchical Generative Framework for Multi-Modal Data Distributions

## Supplementary Material

## Introduction

We present a cluster assignment algorithm which is used in the *GAN-Tree* framework to assign cluster labels to new test samples. We also show further splits of *GAN-Tree* trained on the ImageNet [31] dataset and discuss the flexibility of *GAN-Tree* which allows it to model even outlier clusters. Following this, we show a step by step implementation of the *iGAN-Tree* incremental training algorithm (Algorithm 4 of the main paper). To encourage reproducible research, we will make the code available publicly.

## Unsupervised Classification using *GAN-Tree*

After obtaining a fully-trained *GAN-Tree*, the hierarchical tree framework is used to assign cluster label or object label category for a test sample $x$ satisfying the input training distribution. The entire tree is traversed, starting from the root node to one of the leaf nodes, following the maximum inference probability given by the encoder network $E_{post}^{(n)}(x)$ at each tree node $n$ (see Algorithm 1). The nodes traversed by $x$ at different levels of the *GAN-Tree* is also utilized to assign hierarchical categorization of the sample $x$. This demonstrates the advantage of the inherent top-down divisive clustering setting of *GAN-Tree* as compared to other multi-modal [15; 35] or multi-generator [13; 18] approaches. We achieve 96% accuracy on the MNIST test-set classification in a fully unsupervised setting.

---

**Algorithm 1** *Assign Label* procedure for cluster label assignment of a new sample $x$, given a *GAN-Set* G from *GAN-Tree* T

1: **input:** $x$, G, T
2: $k \leftarrow 0$   ▷ index of the root node of T
3: **while** $k \notin$ G **do**
4:   $z \leftarrow E_{post}^{(k)}(x)$
5:   $k \leftarrow argmax(p(i|z))$   ▷ $i \in children(k)$
6: **return** $k$

---

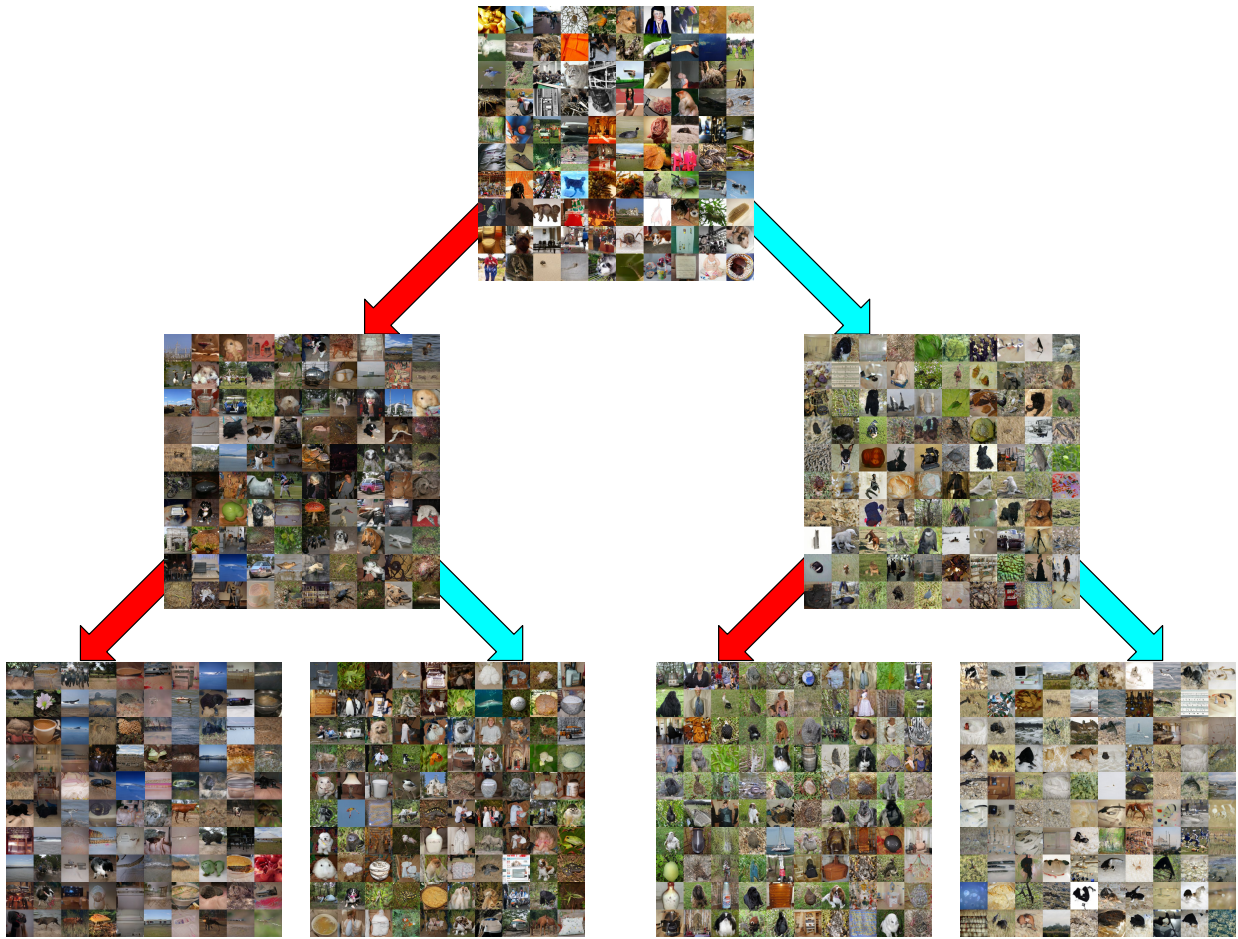

Figure 1: Splits of *GAN-Tree* trained on ImageNet.

## Further Splits of *GAN-Tree* Trained on ImageNet

Figure 1 shows the splits obtained by *GAN-Tree* trained on the ImageNet [31] dataset. The root node is split into images portraying scenery or having objects on a darker background (left child node) and into images portraying greenery or objects on a lighter background (right child node). A further split of the left child node of the root is primarily into images with scenery (left) and those with animals (right). A further split of the right child node of the root is into images with greenery (left) and those with objects on a light-colored background (right).

## Flexibility in Prior Probability Assignment

The prior probabilities defined for each node $n$ can be updated manually depending on the requirement. For instance, consider a curated MNIST dataset with an equal number of samples for all digit classes except a single class consisting of 50% fewer samples as compared to others. In such

a case, the proposed *GAN-Tree* framework has a leaf node dedicated to that particular class with a reduced number of samples. However, during generation, samples are taken from the leaf nodes with uniformly distributed prior probabilities, even in the presence of data imbalance in the actually given training set. We observe decorated quality of generation for samples of that particular class despite the unavailability of many samples to approximate the corresponding distribution. This demonstrates the effectiveness of *GAN-Tree* to model outlier clusters in data distribution even with very few numbers of input samples.

## Step-by-Step Implementation of *iGAN-Tree*

We show a step by step implementation of the *iGAN-Tree* Incremental Training Algorithm (Algorithm 4 of the main paper, copied to Algorithm 2 for reference) for adding the digit class 5 to a *GAN-Tree* trained on MNIST digit classes 0-4. The procedure is portrayed in Figure 2.

- The Base *GAN-Tree* $T$, which has been trained on MNIST digit classes 0 to 4, is given along with the new Data Sample set $\mathcal{D}'$, which consists of the MNIST digit class 5 (see Fig. 2A).

---
**Algorithm 2** Incremental *GAN-Tree* Training
---
1: **input:** *GAN-Tree* $T$, New Data Sample set $\mathcal{D}'$
2: $i \leftarrow \text{index}(\text{root}(T))$
3: **while** $i$ is NOT a leaf node **do**
4:     $l \leftarrow \text{left}(i)$; $r \leftarrow \text{right}(i)$
5:     **if** $\text{Avg}(p_x^{(l)}(\mathcal{D}')) \geq p_z^{(l)}(\mu^{(l)} + d_{\sigma_0})$ **then** $i \leftarrow l$
6:     **else if** $\text{Avg}(p_x^{(r)}(\mathcal{D}')) \geq p_z^{(r)}(\mu^{(r)} + d_{\sigma_0})$ **then** $i \leftarrow r$
7:     **else** break
8: Here, $i$ is the current node index
9: $j \leftarrow \text{NewId}()$ (new parent index)
10: $k \leftarrow \text{NewId}()$ (new child index)
11: $\text{par}(k) \leftarrow j$; $\text{par}(j) \leftarrow \text{par}(i)$; $\text{par}(i) \leftarrow j$
12: **if** $i = \text{index}(\text{root}(T))$ **then**
13:     $\text{root}(T) \leftarrow j$; $\text{left}(j) \leftarrow i$; $\text{right}(j) \leftarrow k$
14: **else if** $i$ was the left child of its previous parent **then**
15:     $\text{left}(\text{par}(j)) \leftarrow j$; $\text{left}(j) \leftarrow i$; $\text{right}(j) \leftarrow k$
16: **else**
17:     $\text{right}(\text{par}(j)) \leftarrow j$; $\text{left}(j) \leftarrow k$; $\text{right}(j) \leftarrow i$
18: Create networks $G^{(k)}, D^{(k)}$ with random initialization
19: Train $E^{(j)}, G^{(k)}$ and $D^{(k)}$ with $\mathcal{L}_{recon}$ and $\mathcal{L}_{adv}$
20: $E^{(par(j))} \leftarrow \text{copy}(E^{(par(i))})$
21: $G^{(j)} \leftarrow \text{copy}(G^{(i)})$; $D^{(j)} \leftarrow \text{copy}(D^{(i)})$; $i \leftarrow \text{par}(i)$
22: **while** $GN^{(i)}$ is not $\text{root}(T)$ **do**
23:     IncrementalNodeTrain$(i)$; $i \leftarrow \text{par}(i)$
    **end while**
24: Train $GN^{(i)}$ with GAN Training Procedure on $\mathcal{D}'$ and generated samples from Terminal GAN-Set.
---

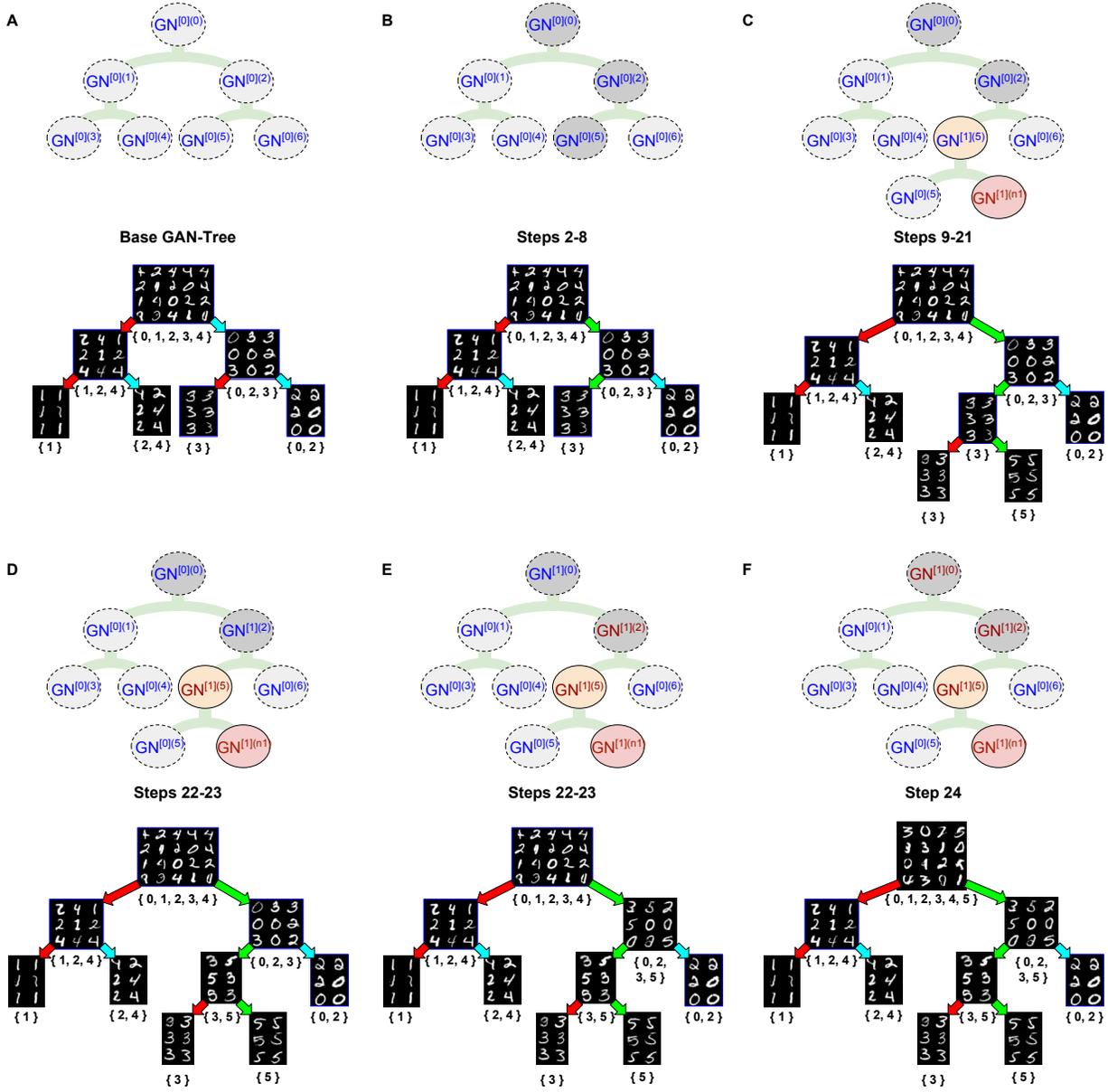

Figure 2: Incremental GAN-Tree training algorithm procedure for a Base *GAN-Tree T* trained on MNIST digit classes 0-4 and new Data Sample $\mathcal{D}$' which consists of MNIST digit class 5.

- The procedure for seeking the apt position to insert the new child node is run in a top-down fashion (lines 2-8 of Algorithm 2). It starts with the root node and checks which child node of the current node best models the data samples from $\mathcal{D}$'. After selecting a child node, this step is recursed on this selected node until a leaf node is reached or its children cannot model the new data samples appropriately. The position at which this seek procedure stops is where the new child node is to be inserted. The seek procedure path ends at $GN^{[0](5)}$ for the case taken, and is highlighted in dark gray (see Fig. 2B).

- Once the position at which the new child node is to be inserted is found, two new nodes are created (lines 9-11 in Algorithm 2): i) a parent node with ID $j$ ($GN^{[1](5)}$ in the case taken), and ii) a child node with ID $k$ which will model only $\mathcal{D}$' ($GN^{[1](n1)}$ in the case taken). The node at which the seek procedure ends ($GN^{[0](5)}$ for the case taken) becomes the child of the new parent node, according to whichever child it was of its previous parent, so as to maintain the same prior distribution (lines 12-17 in Algorithm 2). The child node is then trained after which the new parent node is assigned weights (lines 18-21 in Algorithm 2). The new parent node is highlighted in orange, the new child node in red, and the trained nodes in red text (see Fig. 2C).

- Once the new child node has been trained, the ancestor nodes are trained recursively (lines 22-23 of Algorithm 2), according to the Incremental Node Training Algorithm (given in Algorithm 3 of main paper). The recursive steps are shown in Fig. 2D and Fig. 2E. The node after retraining is highlighted in red text.

- Finally, the root is retrained with the GAN Training Procedure on samples from $\mathcal{D}$' and the samples generated from the GAN-Set formed by the leaf nodes (line 24 in Algorithm 2). The *GAN-Tree* structure after the root node training is shown in Fig. 2F.



\newpage
{\small
\bibliographystyle{ieee_fullname}
\bibliography{egbib}
}

\end{document}